\documentclass{article}

\PassOptionsToPackage{numbers, sort&compress}{natbib}

\usepackage[preprint]{neurips_2023}

\usepackage[utf8]{inputenc} 
\usepackage[T1]{fontenc}    
\usepackage{hyperref}       
\usepackage{url}            
\usepackage{booktabs}       
\usepackage{amsfonts}       
\usepackage{nicefrac}       
\usepackage{microtype}      
\usepackage{xcolor}         

\hypersetup{
    colorlinks=true,
}

\usepackage{comments}
\usepackage{graphicx}
\usepackage{subfigure}
\usepackage{amsmath,amssymb,amsfonts,mathtools,amsthm}
\usepackage[bibliography=separate,forwardlinking=no]{apxproof}
\usepackage[capitalize,noabbrev]{cleveref}

\usepackage{mleftright}
\usepackage{overpic}
\usepackage{pgfplots}
\pgfplotsset{compat=1.18}
\usepgfplotslibrary{fillbetween}
\usepackage{tikz}
\usetikzlibrary{shapes.geometric,decorations.pathreplacing,arrows,patterns}

\tikzstyle{alg} = [rectangle, minimum width=1.3cm, minimum height=0.8cm, text centered, text width=1cm, draw=black]
\tikzstyle{io} = [rectangle, minimum width=3cm, minimum height=1cm, text centered, text width=3cm, draw=black]
\tikzstyle{nn} = [rectangle, minimum width=3cm, minimum height=0.5cm, text centered, text width=6cm, draw=black]
\tikzstyle{arrow} = [thick,->,>=stealth]

\theoremstyle{plain}
\newtheoremrep{theorem}{Theorem}[section]
\newtheoremrep{proposition}[theorem]{Proposition}
\newtheoremrep{lemma}[theorem]{Lemma}
\newtheoremrep{corollary}[theorem]{Corollary}
\theoremstyle{definition}
\newtheoremrep{definition}[theorem]{Definition}
\newtheoremrep{assumption}[theorem]{Assumption}
\theoremstyle{remark}
\newtheoremrep{remark}[theorem]{Remark}

\graphicspath{{Figures/}}

\DeclareMathOperator*{\argmax}{arg\,max}

\def\D{D(\theta_s,\theta_q)}
\def\J{J(\theta_s,\theta_q)}
\newcommand{\p}[1]{\mathbb{P}\mleft(#1\mright)}
\def\corr{\mathrm{Corr}}
\def\cov{\mathrm{Cov}}

\makeatletter
\newcommand*{\e}{%
  \def\e@sub{}%
  \e@scripts
}
\newcommand*{\e@scripts}{%
  \@ifnextchar_\e@subscript{%
    \e@finish
  }%
}
\def\e@subscript_#1{%
  \ifx\e@sub\@empty
    \def\e@sub{#1}%
  \else
    \errmessage{e already has a subscript}%
  \fi
  \e@scripts
}
\newcommand*{\e@finish}[1]{%
  \mathbb{E}%
  \ifx\e@sub\@empty\else _{\e@sub}\fi
  \mleft[#1\mright]%
}
\makeatother

\title{Coincident Learning for Unsupervised Anomaly Detection}

\author{
    Ryan Humble$^1$\thanks{Correspondence to ryhumble@stanford.edu.}, Zhe Zhang$^2$, Finn O'Shea$^2$, Eric Darve$^{1,3}$, Daniel Ratner$^2$ \\
    $^1$ Institute for Comptuational and Mathematical Engineering, Stanford University \\
    $^2$ SLAC National Laboratory \\
    $^3$ Department of Mechanical Engineering, Stanford University
}

\begin{document}

\maketitle

\begin{abstract}
Anomaly detection is an important task for complex systems (e.g., industrial facilities, manufacturing, large-scale science experiments), where failures in a sub-system can lead to low yield, faulty products, or even damage to components. While complex systems often have a wealth of data, labeled anomalies are typically rare (or even nonexistent) and expensive to acquire. Unsupervised approaches are therefore common and typically search for anomalies either by distance or density of examples in the input feature space (or some associated low-dimensional representation). This paper presents a novel approach called CoAD, which is specifically designed for multi-modal tasks and identifies anomalies based on \textit{coincident} behavior across two different slices of the feature space. We define an \textit{unsupervised} metric, $\hat{F}_\beta$, out of analogy to the supervised classification $F_\beta$ statistic. CoAD uses $\hat{F}_\beta$ to train an anomaly detection algorithm on \textit{unlabeled data}, based on the expectation that anomalous behavior in one feature slice is coincident with anomalous behavior in the other. 
The method is illustrated using a synthetic outlier data set and a MNIST-based image data set, and is compared to prior state-of-the-art on two real-world tasks: a metal milling data set and a data set from a particle accelerator.
\end{abstract}

\section{Introduction}

The problem of anomaly detection, the task of finding abnormal events or data, is an important task for complex systems, such as industrial facilities, manufacturing, and large-scale science experiments~\cite{Sun16,
Zhao19,Lutz20,Edelen21,Lindemann21,Radaideh22}. Failures in these systems can lead to low yield, faulty products, or even damage to components, making identifying these failures a high-priority task for system operators. However, the complexity of these systems typically ensures that labeled data is rare or nonexistent and expensive to acquire. 

This makes traditional AD methods challenging to apply. While many AD approaches exist, they can generally be categorized into three broad classes. Probability-based methods~\cite{reynolds2009gaussian,pena2001multivariate,hubert2010minimum}, such as Gaussian mixtures, are mathematically straightforward to analyze but are restricted to datasets with simple geometric shapes. Distance-based~\cite{kramer2013k,kriegel2008angle,li2003improving,Liu08,Scholkopf01} and density-based methods~\cite{alghushairy2020review,he2003discovering,papadimitriou2003loci,janssens2012stochastic,Breunig00,Parzen62}, such as LOF, assume that anomalies are well-separated from nearby clusters of normal samples, but they are less accurate when dealing with noisy features and when the training set contains many anomalies that cluster together, which is a common occurrence in scientific datasets. Compression-based techniques, such as kernel PCA~\cite{hoffmann2007kernel} and autoencoders~\cite{zenati2018adversarially,yang2020regularized,schlegl2019f,liu2019generative,zhou2021vae}, are effective in some situations, but their performance deteriorates when anomalous samples are present in the training set. 


\begin{figure}[htbp]
    \centering
    \begin{minipage}{.56\textwidth}
    \includegraphics[width=\linewidth]{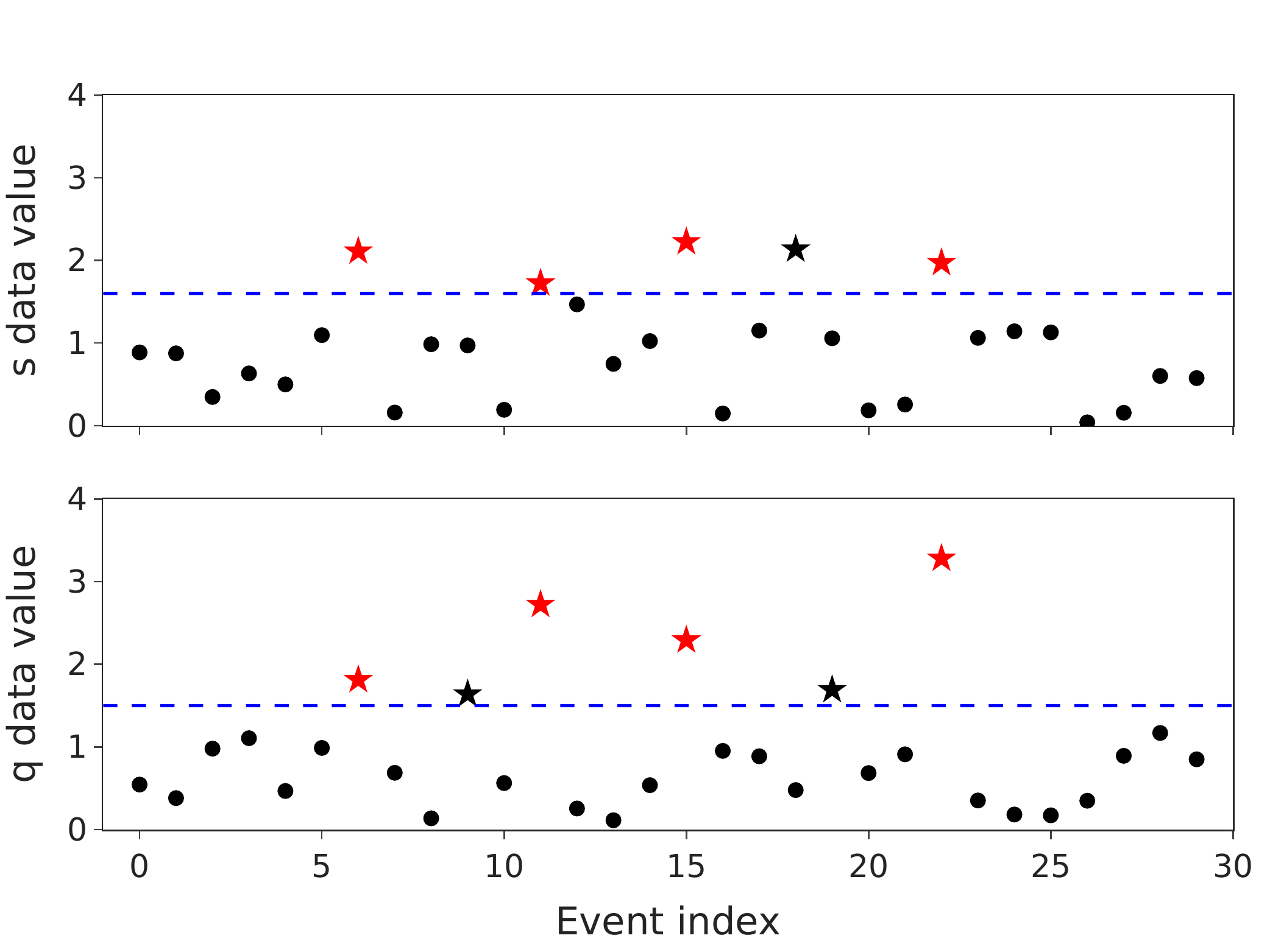}
    \end{minipage}\hspace{1em}
    \begin{minipage}{.38\textwidth}
    \caption{We consider tasks with two inputs, $s$ and $q$, which should be independent except for anomalous events which exist across both inputs. Anomaly detection models identify points independently in each input (stars) using example thresholds (dashed blue line). Events found in only one input (black stars) are ignored, and only joint events found in both inputs are identified as anomalous (red stars).
    }\label{fig:synth_data}
    \end{minipage}
\end{figure}

This work presents a novel approach based on coincidence learning between two data inputs $ s $ and $ q $. By relying on the expectation that anomalies affect and occur simultaneously in both inputs, our approach learns to differentiate between normal and anomalous examples. Using this concept of coincidence, we are able to overcome several limitations of existing state-of-the-art methods, including handling training sets with clustered anomalous examples and being resilient to noisy input features. We emphasize that this method can be applied to any scenario where two sets of measurements of a physical system are available, or when a system is composed of a series of stages, each equipped with sensors. Even when only a single measurement is available, the algorithm can still be employed if two data inputs $s$ and $q$ can be generated by splitting the measurement data. 

Our motivating example comes from a particle accelerator, for which we have two data inputs, one containing data from a radio frequency (RF) station subsystem ($s$) and one containing electron energy data beam-position monitors (BPMs) that monitor beam quality ($q$). During normal operation, the variability in the signals is independent (i.e., random fluctuations in $s$ and $q$ are uncorrelated). However, the anomalous behavior of an RF station will have an impact, albeit \textit{unknown}, on the BPM quality data. Even with no ground truth labels for training, we will show that we can exploit the coincidence of abnormalities to determine whether abnormal RF station subsystem behavior has caused beam performance degradation.

\subsection{Contributions}
In this paper, we consider a subset of anomaly detection tasks in which the data is partitioned into two inputs $s $ and $q$. \cref{fig:synth_data} demonstrates our problem setting. We assume that the impact of an anomaly is apparent in both inputs. As a consequence, we expect that there exists an algorithm capable of dividing each input into normal and anomalous sets, and crucially the sets should match. Additionally, we assume that $s$ and $q$ data are independent within either the normal or anomalous clusters. 

Our main contributions are:
\begin{enumerate}
    \item We introduce coincident learning for anomaly detection (CoAD) and an unsupervised metric $ \hat{F}_\beta $, in analogy to the supervised classification metric $ F_\beta $, that exploits the coincidence between $ s $ and $ q $ to classify normal and anomalous examples. Specifically, we use two models---that use different data inputs---to classify examples as either normal or anomalous.

    \item We present theoretical results showing that our estimate $ \hat{F}_\beta $ is a lower bound of the true $ F_\beta $ under our assumptions and derive the form of the optimal models under mild conditions.

    \item We show that $ \hat{F}_\beta $ can be used in two modes: categorical and continuous. If the two anomaly detection models are predefined, $ \hat{F}_\beta $ gives a principled way of selecting the two thresholds (\cref{sec:thresh_results}). We can also use $ \hat{F}_\beta $ to train the anomaly models end-to-end, with the models parameterized as deep neural networks (DNNs) (\cref{sec:mnist,sec:milling,sec:particle_accel}).

    \item We demonstrate these contributions on four data sets: a synthetic outlier data set, a synthetic image data set generated from MNIST, a publicly available metal milling data set, and an experimental data set taken from a particle accelerator. For the synthetic cases, we show our unsupervised method performs nearly as well as a supervised counterpart. For the real data sets, we train DNNs end-to-end to achieve data-driven, unlabeled anomaly detection and show improvements over prior state-of-the-art.
\end{enumerate}

\section{Coincident Learning} \label{sec:coin_ad}
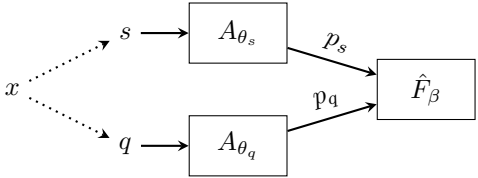
\begin{figure}[htbp]
    \centering
    \begin{minipage}{.5\textwidth}
    \begin{tikzpicture}[node distance=1.5cm]
        \node (hidden) [] {$x$};
        \node (s_in) [right of=hidden,yshift=0.75cm] {$s$};
        \node (q_in) [right of=hidden,yshift=-0.75cm] {$q$};
        \node (A_s) [alg,right of=s_in] {$A_{\theta_s}$};
        \node (A_q) [alg,below of=A_s] {$A_{\theta_q}$};
        \node (method) [alg,right of=A_s, xshift=1cm, yshift=-0.75cm] {$\hat{F}_\beta$};

        \draw [arrow,dotted] (hidden) -- (s_in);
        \draw [arrow,dotted] (hidden) -- (q_in);
        \draw [arrow] (s_in) -- (A_s);
        \draw [arrow] (q_in) -- (A_q);
        \draw [arrow] (A_s) -- (method) node [midway, above, sloped] {$p_s$};
        \draw [arrow] (A_q) -- (method) node [midway, above, sloped] {$p_q$};
    \end{tikzpicture}        
    \end{minipage}\hspace{1em}
    \begin{minipage}{.46\textwidth}
    \caption{A schematic of CoAD showing the two data inputs $s$ and $q$ (generated from a hidden state $x$) and their respective models $A_{\theta_s}$ and $A_{\theta_q}$, which are trained to maximize our unsupervised metric $\hat{F}_\beta$.}
    \label{fig:schem}
    \end{minipage}
\end{figure}

Given a dataset \( \mathcal{D} = \{(s,q)\} \) drawn from the respective data inputs, we consider a pair of models, $A_{\theta_s}(s)$ and $A_{\theta_q}(q)$, parameterized by $\theta_s$ and $\theta_q$. The models will each have a scalar output, $p_s, p_q \in [0,1]$, which we will interpret as the confidence that the example belongs to the anomalous class. Let \( \mathcal{D}_s = \{s\} \) and \( \mathcal{D}_q = \{q\} \) be the marginal datasets for $ s $ and $ q $ respectively. Also, we assume that the data is generated from an unseen state variable $ x $ (i.e., both are functions of x: $ s(x) $ and $ q(x) $). We present a schematic of our method in~\cref{fig:schem}.

\subsection{CoAD Objective}
\begin{toappendix}
    \subsection{CoAD Objective (Proofs)}
\end{toappendix}

To begin, we restrict ourselves to the case where $p_s$ and $p_q$ are categorical labels in $\{0,1\}$. We define a \textit{joint event} as one where both models classify the respective data as anomalous, i.e., \( p_{s,i} = p_{q,i} = 1 \). Since we lack true labels, we cannot determine which joint events are true positives (true anomalous events) or are false positives (normal events flagged as anomalies). However, we show below that we can estimate the number of false positives from the disagreement between the two models. We can then compare the actual number of observed joint events with the estimated number of false positive events to measure the efficacy of our algorithm. The more joint events we observe, the more sensitive our algorithm. The fewer points with conflicting predictions (and thus fewer estimated false positives), the more precise our algorithm.

Let $ \J $ denote the fraction of joint events found in our data (i.e., predicted positives). Suppose $ \alpha $ is the anomaly fraction in our data (i.e., actual positives). If we had labels for each of the $ n $ examples in our dataset, we would evaluate our algorithm with the supervised classification metric \( F_\beta \); however, since we lack labels, we propose an unsupervised version \( \hat{F}_\beta \) that relies on an estimate $ \D $ of the fraction of false positives. That is, we approximate
\begin{align}
    F_\beta = \frac{(1 + \beta^2) (J - \text{FP}/n)}{J + \alpha \beta^2} \qquad\Longrightarrow\qquad \hat{F}_\beta = \frac{(1 + \beta^2) (J-D)}{J + \alpha \beta^2}, \label{eqn:fbeta_compare}
\end{align}
where higher values are better, FP is the number of false positives, and $\beta$ balances the weighting of precision and recall. (We also show later an estimate of $ \alpha $ suffices.)

The quantity $\hat{F}_\beta$ can now be used to compare algorithms or select model hyperparameters in the same manner as its supervised counterpart. As in the supervised case, the extremes are $ \hat{P} = \hat{F}_0 $ and $ \hat{R} = \hat{F}_\infty $, which correspond to precision \( P \) and recall \( R \). We can use $\hat{F}_\beta$ to pick a model that strikes a balance between the number of anomalous events found (maximizing the recall) and the confidence in the prediction (maximizing precision).

The \( \hat{F}_\beta \) definition requires an estimate of the fraction of false positives \( \D \). Our strategy is based on an observation that \textit{disagreements} between $ A_{\theta_s} $ and $ A_{\theta_q} $ reveal the false positive rate of our algorithm. Under the assumption that $ s $ and $ q $ are independent conditioned on knowing the true label, \cref{thm:disagree_rate} shows that the disagreement rates provide an upper bound on the true fraction of false positives. (All proofs are deferred to the Appendix.)
\begin{theoremrep}\label{thm:disagree_rate}
    Assume that $s$ and $q$ are independent conditioned on knowing the true label, and assume that the models \( A_{\theta_s}, A_{\theta_q} \) are no worse than random guessers. Define $ \D = \e_{(s,q) \in \mathcal{D}}{p_s|\neg p_q} \e_{(s,q) \in \mathcal{D}}{p_q | \neg p_s} $ to be our estimated fraction of false positives. Then, in the categorical case, the fraction of false positives is no more than $ D $ and the fraction of true positives is at least $J - D$, implying that $\hat{R}, \hat{P} $, and $ \hat{F}_\beta $ are lower bounds of their supervised counterparts $ R, P $, and $ F_\beta$.
\end{theoremrep}
\begin{proof}
    Let $ y(s,q) $ denote the true label ($0$-normal, $1$-anomalous). For categorical \( p_s(s), p_q(q) \), the fraction of false positives is
    \begin{align*}
        \e_{\mathcal{D}}{p_s p_q \neg y} & = \e{p_s | \neg y} \e{p_q | \neg y} \p{\neg y}
    \end{align*}
    where we use the independence of $ s $ and $ q $ conditioned on the true label. Since we do not have the labels, we cannot directly measure \( \e{p_s | \neg y} \) and \( \e{p_q | \neg y} \). We instead will bound this expectation:
    \begin{align*}
        \e{p_s | \neg y} = \p{p_s | \neg y} &= \p{p_s | \neg p_q, \neg y} \\
        & = \p{p_s | \neg p_q} - \left( \p{p_s | y} - \p{p_s | \neg y} \right) \p{y | \neg p_q} \\
        & \leq \p{p_s | \neg p_q} = \e{p_s | \neg p_q}
    \end{align*}
    where in the first line we use the independence of $s$ and $q$ within a single class of data, and in the final line we use $\p{p_s|y} \geq \p{p_s|\neg y}$ (i.e., the model is no worse than random). In the second line, we use the identity
    \begin{align*}
        \p{p_s|\neg p_q} &= \p{p_s|\neg p_q,\neg y} \p{\neg y | \neg p_q} + \p{p_s | \neg p_q, y} \p{y | \neg p_q} \\
        &= \p{p_s | \neg p_q, \neg y} \left( 1 - \p{y| \neg p_q}\right) + \p{p_s | \neg p_q, y} \p{y | \neg p_q } \\
        &= \p{p_s | \neg p_q, \neg y} + \left[ \p{p_s | \neg p_q, y} - \p{p_s | \neg p_q, \neg y} \right] \p{y | \neg p_q} \\
        &= \p{p_s | \neg p_q, \neg y} + [\p{p_s|y} - \p{p_s|\neg y}] \p{y | \neg p_q}
    \end{align*}
    where we have again applied the independence of $s$ and $q$ within a single class. Finally plugging in $\e{p_s | \neg y} \leq \e{p_s | \neg p_q}$, $\e{p_q | \neg y} \leq \e{p_q | \neg p_s} $ (by the same logic), and $ \p{\neg y}\leq 1$, we find
    \begin{align*}
        \e_{\mathcal{D}}{p_s p_q \neg y} \leq \e{p_s | \neg p_q} \e{p_q | \neg p_s} = \D
    \end{align*}
    Lastly, since \( \e_{\mathcal{D}}{p_s p_q \neg y} \leq \D \), the fraction of true positives is
    \begin{align*}
        \e_{\mathcal{D}}{p_s p_q y} & = \e_{\mathcal{D}}{p_s p_q} - \e_{\mathcal{D}}{p_s p_q y} \\
        & \geq \J - \D
    \end{align*}
    and therefore \( \hat{F}_\beta \leq F_\beta \).
\end{proof}
\begin{corollaryrep}\label{thm:naive_rate}
    Additionally define \( D_\text{naive}(\theta_s, \theta_q) = \e_{s \in \mathcal{D}_s}{p_s} \e_{q \in \mathcal{D}_q}{p_q} \), which is equivalent to an assumption that $ s $ and $ q $ are completely independent. Then, \( \D \leq D_\text{naive} \).
\end{corollaryrep}
\begin{proof}
    We have
    \begin{align*}
        \e_{\mathcal{D}}{p_s} & = \e{p_s | p_q} \p{p_q} + \e{p_s | \neg p_q} \p{\neg p_q} \\
        & = \e{p_s | \neg p_q} + \p{p_q} \left( \e{p_s | p_q} - \e{p_s | \neg p_q} \right) \\
        & \geq \e{p_s | \neg p_q}
    \end{align*}
    since \( \e{p_s | p_q} \geq \e{p_s | \neg p_q} \) if the models are no worse than random.
\end{proof}
Therefore, by definition of a joint event and the conditional expectation, the fraction of joint events $J$ in the data and the estimated fraction of false positives $D$ are
\begin{equation}
    \J = \mu_{sq}, \qquad 
    \D = \frac{\mu_s - \mu_{sq}}{1 - \mu_q} \frac{\mu_q - \mu_{sq}}{1 - \mu_s}, 
\end{equation}
where \( \mu_s = \e_{s \in \mathcal{D}_s}{p_s} \), \( \mu_q \) similarly, and \( \mu_{sq} = \e_{(s,q) \in \mathcal{D}}{p_s p_q} \).
This allows us to concretely write our unsupervised metric as
\begin{align}
    \hat{F}_\beta & = (1 + \beta^2) \frac{\mu_{sq} - \mu_s \mu_q}{\mu_{sq} + \alpha \beta^2} \frac{1 - \mu_{sq}}{(1 - \mu_s) (1 - \mu_q)}.
\end{align}
It is crucial to note that we implicitly require a majority of the events to be labeled as \( 0 \), or equivalently require the anomalous class to be the minority class; this is a necessary condition since precision and recall (and our unsupervised analogues) are not invariant under a labeling flip. Thus, we additionally impose the constraint that anomalies exist and are rare (\( 0 < \mu_{sq} \leq \mu_s, \mu_q \leq 0.5 \)). Also, as a point of optimization, it might appear that maximizing \( \hat{F}_\beta \) requires that both \( \alpha \) and \( \beta \) be defined. However, as \( 1 + \beta^2 \) is just a constant scalar, we need only specify the quantity \( \alpha \beta^2 \).
If we used an incorrect estimate of \( \alpha \) (since the true \( \alpha \) is unknown), we have merely maximized our metric for a different value of \( \beta \). Moreover, the maximizers of \( \hat{P} \) and \( \hat{R} \) do not depend on \( \alpha \) at all.

Lastly, when developing our metric, we assumed the categorical case (\( p_s, p_q \in \{0,1\} \)). This case might naturally arise when the two models \( A_{\theta_s}, A_{\theta_q} \) already exist and are parameterized by two thresholds. The metric then allows a principled way of setting these thresholds, as we demonstrate in \cref{sec:thresh_results}. But notably, our metric naturally extends to the case of continuous $p_s, p_q \in [0,1]$. This allows us to train more complex models \( A_{\theta_s}, A_{\theta_q} \), such as ones parameterized as DNNs and trained with gradient-based optimizers, thereby allowing us to cluster normal and anomalous data without having first to build the individual anomaly detection models. We present a theoretical justification for the continuous extension in the following section.


\subsection{Properties of CoAD}
\begin{toappendix}
    \subsection{Properties of CoAD (Proofs)}
\end{toappendix}

Under certain simplifying assumptions, we can derive results regarding the solution to the optimization problem. Throughout we assume that \( p_s(s) = A_{\theta_s}(s) \), for some choice of parameters \( \theta_s \), can map \( D_s \) to any element of \( [0,1]^n \) where $n$ is the number of samples in the training set (and similarly for \( p_q(q) = A_{\theta_q}(q) \)).

We first show that the maximizers $ p_s^*, p_q^*$ of $ \hat{F}_\beta $ are (nearly) categorical solutions. Therefore, the continuous extension (to $ p_s, p_q \in [0,1] $) is almost equivalent to the original categorical case, and our method behaves as a (nearly) hard clustering algorithm. Concretely, holding \( p_s \) fixed, \cref{thm:near_cat} shows that the optimal $ p_q^* $ is (nearly) categorical: \( p_q^*(q) \in \{ 0, \rho, 1 \} \) for some \( \rho \in [0,1] \) and all $ q $. Moreover, \cref{thm:cat_tightness} shows \( \rho \not\in \{0,1\} \) only occurs if the constraint \( \mu_q \leq 0.5 \) is tight. By applying this twice (first for fixed $ p_s $ and then again with the new $p_q$ fixed), we need only consider (nearly) categorical solutions for $ p_s, p_q $. 
\begin{theoremrep}\label{thm:near_cat}
    Assume $ \theta_s $ is fixed (with \( \mu_s \in (0, 0.5] \)). Let $ w(q) = \e_{s|q \in \mathcal{D}_{s|q}}{p_s(s)} $. Then, the maximum of $ \hat{F}_\beta $ can be achieved by a (nearly) categorical solution: $ p_q^*(q) = \mathbf{1}\left\{w(q) > \tau\right\} + \rho \mathbf{1}\left\{w(q) = \tau\right\}$ for some \( \rho, \tau \in [0,1] \).
\end{theoremrep}
\begin{proof}
    We can directly optimize as follows
    \begin{align*}
        \max_{\substack{p_q \in [0,1]^n\\ \mu_q \in [0, 0.5]}} \hat{F}_\beta & = (1 + \beta^2) \max_{\substack{p_q \in [0,1]^n\\ \mu_q \in [0, 0.5]}} \frac{\mu_{sq} - \mu_s \mu_q}{\mu_{sq} + \beta^2 \alpha} \frac{1 - \mu_{sq}}{(1 - \mu_s) (1 - \mu_q)} \\
        & = \frac{1 + \beta^2}{1 - \mu_s} \left( \max_{\substack{p_q \in [0,1]^n\\ \mu_q \in [0, 0.5]}} \frac{\mu_{sq} - \mu_s \mu_q}{\mu_{sq} + \beta^2 \alpha} \frac{1 - \mu_{sq}}{1 - \mu_q} \right) \\
        & = \frac{1 + \beta^2}{1 - \mu_s}  \left[ \max_{\gamma \in [0, 0.5]} \left( \max_{\substack{p_q \in [0,1]^n\\ \mu_q = \gamma}} \frac{\mu_{sq} - \mu_s \gamma}{\mu_{sq} + \beta^2 \alpha} (1 - \mu_{sq}) \right) \frac{1}{1 - \gamma} \right] \\
        & = \frac{1 + \beta^2}{1 - \mu_s}  \left[ \max_{\gamma \in [0, 0.5]} \frac{\mu_{sq}^*(\gamma) - \mu_s \gamma}{\mu_{sq}^*(\gamma) + \beta^2 \alpha} \frac{1 - \mu_{sq}^*(\gamma)}{1 - \gamma} \right]
    \end{align*}
    where \( \mu_{sq}^*(\gamma) \) is the solution to the inner problem. The solution to the unconstrained inner problem is
    \begin{align*}
        \tilde{\mu}_{sq}(\gamma) & = \argmax_{\mu_{sq}} \frac{\mu_{sq} - \mu_s \gamma}{\mu_{sq} + \beta^2 \alpha} (1 - \mu_{sq}) \\
        & = \sqrt{(1 + \alpha \beta^2) (\mu_s \gamma + \alpha \beta^2)} - \alpha \beta^2
    \end{align*}
    since
    \begin{align*}
        \frac{\partial}{\partial \mu_{sq}} \left( \frac{\mu_{sq} - \mu_s \gamma}{\mu_{sq} + \beta^2 \alpha} (1 - \mu_{sq}) \right) & = \frac{(1 + \alpha \beta^2) (\mu_s \gamma +  \alpha \beta^2)}{(\mu_{sq} + \alpha \beta^2)^2} - 1
    \end{align*}
    and
    \begin{align*}
        \frac{\partial^2}{\partial \mu_{sq}^2} \left( \frac{\mu_{sq} - \mu_s \gamma}{\mu_{sq} + \beta^2 \alpha} (1 - \mu_{sq}) \right) & = -2 \frac{(1 + \alpha \beta^2) (\mu_s \gamma +  \alpha \beta^2)}{(\mu_{sq} + \alpha \beta^2)^3} \leq 0.
    \end{align*}

    Now we will show that \( \mu_{sq}^*(\gamma) \leq \tilde{\mu}_{sq}(\gamma) \). First, we have
    \begin{align*}
        \mu_{sq}^2 & = \e{p_s p_q}^2 \\
        & \leq \e{p_s^2} \e{p_q^2} \quad \text{by Cauchy-Schwartz} \\
        & \leq \e{p_s} \e{p_q} \quad \text{since } p_s, p_q \in [0,1] \\
        & = \mu_s \gamma
    \end{align*}
    and
    \begin{align*}
        \mu_{sq} \leq \frac{1}{2} \le \frac{1}{2} (1 + \mu_s \gamma) \quad \text{since } \mu_{sq} \leq \mu_q \leq \frac{1}{2}.
    \end{align*}
    We can now bound $(\mu_{sq} + \alpha \beta^2)^2$:
    \begin{align*}
        (\mu_{sq} + \alpha \beta^2)^2 & = \mu_{sq}^2 + 2 \mu_{sq} \alpha \beta^2 + (\alpha \beta^2)^2 \\
        & \leq \mu_s \gamma + (1 + \mu_s \gamma) \alpha \beta^2 + (\alpha \beta^2)^2 \\
        & = (1 + \alpha \beta^2) (\mu_s \gamma + \alpha \beta^2) \\
        & = \left( \tilde{\mu}_{sq}(\gamma) + \alpha \beta^2 \right)^2.
    \end{align*}
    Therefore, we have shown \( \mu_{sq}^*(\gamma) \leq \tilde{\mu}_{sq}(\gamma) \), and thus the (constrained) inner optimization problem is concave increasing in \( \mu_{sq} \) and amounts to just maximizing $ \mu_{sq} $:
    \begin{align*}
        \mu_{sq}^*(\gamma) & = \argmax_{\mu_{sq}: \substack{p_q \in [0,1]^n\\ \mu_q = \gamma}} \mu_{sq} \\
        & = \argmax_{\mu_{sq}: \substack{p_q \in [0,1]^n\\ \mu_q = \gamma}} \e_{q \in \mathcal{D}_q}{p_q(q) \e_{s|q \in \mathcal{D}_{s|q}}{p_s(s)}} \\
        & = \argmax_{\mu_{sq}: \substack{p_q \in [0,1]^n\\ \mu_q = \gamma}} \e_{q \in \mathcal{D}_q}{p_q(q) w(q)}.
    \end{align*}
    In order to maximize this, we first assign \( p_q^* \) mass to those \( q \) with largest \( w(q) \) and then progressively to those with smaller \( w(q) \). We therefore have
    \[
        p_q^*(q) = \begin{cases}
            1 & w(q) > \tau(\gamma) \\
            \rho(\gamma) & w(q) = \tau(\gamma) \\
            0 & w(q) < \tau(\gamma)
        \end{cases}
    \]
    where \( \rho(\gamma), \tau(\gamma) \) are set to achieve \( \mu_q = \gamma \). Specifically, let \( f_{w(q)}(w) = \p{w(q) = w} \) and \( \bar{F}_{w(q)}(w) = \p{w(q) > w} \). If \( \exists \tau(\gamma) \) such that \( \bar{F}_{w(q)}(\tau(\gamma)) = \gamma \), then \( \rho(\gamma) = 0 \). Otherwise, let $ \tau(\gamma) = \inf \{ w: \bar{F}_{w(q)}(w) \leq \gamma \} $. By definition, this implies that \( f_{w(q)}(\tau(\gamma)) \geq \gamma - \bar{F}_{w(q)}(\tau(\gamma)) \), so we can set $ \rho(\gamma) = \frac{\gamma - \bar{F}_{w(q)}(\tau(\gamma))}{f_{w(q)}(\tau(\gamma))} \in [0,1] $. Generally, we have
    \begin{align*}
        \tau(\gamma) & = \inf \{ w: \bar{F}_{w(q)}(w) \leq \gamma \} \\
        \rho(\gamma) & = \begin{cases}
            0 & \text{if } \bar{F}_{w(q)}(\tau(\gamma)) = \gamma \\
            \frac{\gamma - \bar{F}_{w(q)}(\tau(\gamma))}{f_{w(q)}(\tau(\gamma))} & \text{o.w.}
        \end{cases}
    \end{align*}





\end{proof}
\begin{toappendix}
\begin{corollary}\label{thm:cont_concave}
    Define \( \mu_{sq}^*(\gamma) = \argmax_{p_q \in [0,1]^n, \mu_q = \gamma} \mu_{sq} \). Then, $\mu_{sq}^*(\gamma) $ is continuous in \( \gamma \) and concave increasing.
\end{corollary}
\begin{proof}
    From its definition in~\cref{thm:near_cat}, the function $\mu_{sq}^*(\gamma)$ is increasing and continuous. Moreover, as $\gamma$ increases, $\tau(\gamma)$ decreases. As a consequence, $\mu_{sq}^*(\gamma)$ is concave.
\end{proof}
\end{toappendix}
\begin{theoremrep}\label{thm:cat_tightness}
    Additionally, a non-categorical solution (i.e., with \( \rho \not\in \{0,1\} \)) can only be uniquely optimal if the constraint \( \mu_q \leq 0.5 \) is tight.
\end{theoremrep}
\begin{proof}
    Starting at the end of the proof of \cref{thm:near_cat}, we note that \( \tau(\gamma) \) is the generalized inverse of the complementary cumulative distribution function \( \bar{F}_{w(q)}(w) \). If a unique inverse exists, \( \mu_{sq}^*(\gamma) \) is differentiable on \( (0, 1) \); moreover, \( \rho(\gamma) = 0 \) for all \( \gamma \), and we have a categorical solution for \( p_q^*(q) \) achieving the maximum.
     
    If instead a unique inverse does not exist, \( \mu_{sq}^*(\gamma) \) can have piecewise linear segments, where each segment corresponds to a different \( \tau(\gamma) \) threshold and the endpoints of each segment correspond to different categorical solutions. We need to revisit the outer problem (over \( \gamma \)) to establish the form of the solution. As shown in \cref{thm:near_cat,thm:cont_concave}, \( \mu_{sq}^*(\gamma) \) is concave increasing and is piecewise linear (where each segment corresponds to a different \( \tau(\gamma) \) threshold and the endpoints of each segment correspond to different categorical solutions). We also have that
    \begin{align*}
        \mu_s \gamma & = \e{p_s \gamma} \\
        & \leq \max_{\substack{p_q \in [0,1]^n \\ \mu_q = \gamma}} \e{p_s p_q} = \mu_{sq}^*(\gamma) \\
        & \leq \min \left(\e{p_s 1}, \max_{\substack{p_q \in [0,1]^n \\ \mu_q = \gamma}} \e{1 p_q} \right) = \min(\mu_s, \gamma) \\
        & \leq \frac{1}{2}
    \end{align*}
    and \( 0 \leq \tau(\gamma) \leq 1 \).

    \begin{figure}[htbp]
        \centering
        \begin{tikzpicture}
            \begin{axis}[
              width=3in,
              xmin=0,xmax=0.5, xtick={0,0.1,0.2,0.3,0.4,0.5},
              ymin=0,ymax=0.5, ytick={0,0.1,0.2,0.3,0.4,0.5},
              axis lines=middle,
              enlargelimits,
              ticklabel style={fill=white},
              xlabel=$\gamma$,
              ylabel=$\mu_{sq}(\gamma)$,
              clip=true,
              extra x ticks={0.15, 0.27},
              extra x tick style={%
                grid=major,
                major tick length=15pt,
                xtick align=outside
              },
              extra x tick labels={%
                $\gamma_m$,
                $\gamma_M$
              }
            ]

            \addplot[color=black, mark=x] coordinates {
                (0,0)
                (0.15,0.15)
                (0.27,0.24)
                (0.4,0.29)
                (0.45,0.30)
                (1,0.31)
            };
            \addlegendentry{\(\mu_{sq}^*(\gamma)\)}

            \addplot[color=black, dashed] coordinates {
                (0,0.0375)
                (1,0.7875)
            };
            \addlegendentry{\(\tau(\gamma_m) \gamma + \mu_{sq}^*(\gamma_m)\)}


            \path[name path=top] (axis cs:-1,0.5) -- (axis cs:1,0.5);
            \path[name path=axis1] (axis cs:0,1) -- (axis cs:1,1);
            \addplot[gray, opacity=0.4, pattern=north west lines] fill between[of=top and axis1];
           
            \path[name path=right] (axis cs:0.5,-1) -- (axis cs:0.5,0.5);
            \path[name path=axis2] (axis cs:1,0) -- (axis cs:1,0.5);
            \addplot[gray, opacity=0.4, pattern=north west lines] fill between[of=right and axis2];
            \end{axis}
        \end{tikzpicture}
        \caption{Illustration of piecewise linear behavior of $ \mu_{sq}^*(\gamma)$.}\label{fig:pw_linear} 
    \end{figure}

    We now consider where the maximum lies along each piecewise linear segment. Suppose there is a segment on \( [\gamma_m, \gamma_M] \) with \( 0 \leq \gamma_m < \gamma_M \leq 0.5 \). Along this segment, let \( \mu_{sq}^*(\gamma) = a x + b \) where \( a = \tau(\gamma_m) \), \( b = \mu_{sq}^*(\gamma_m) \), and \( x = \gamma - \gamma_m \). We depict this in~\cref{fig:pw_linear}. The maximum along the segment is defined by
    \begin{align*}
        \argmax_{\gamma \in [\gamma_m, \gamma_M]} \hat{F}_\beta(\gamma) & = \argmax_{\gamma \in [\gamma_m, \gamma_M]} \frac{\mu_{sq}^*(\gamma) - \mu_s \gamma}{\mu_{sq}^*(\gamma) + \alpha \beta^2} \frac{1 - \mu_{sq}^*(\gamma)}{1 - \gamma} \\
        & = \gamma_m + \argmax_{x \in [0, \gamma_M - \gamma_m]} \frac{(b - \mu_s \gamma_m) + (a - \mu_s) x}{(b + \alpha \beta^2) + a x} \frac{(1 - b) - ax}{(1 - \gamma_m) - x} \\
        & = \gamma_m + \argmax_{x \in [0, \gamma_M - \gamma_m]} f(x) g(x).
    \end{align*}
    Since \( \mu_s \gamma \leq \mu_{sq}^*(\gamma) \leq \frac{1}{2} \), we immediately have \( \hat{F}_\beta(\gamma) \geq 0 \) for all \( \gamma > 0 \), which implies the null solution, \( \gamma = 0 \), is either not optimal or is not uniquely optimal, so we ignore it here-forward.
    
    For \( z = \alpha \beta^2 (a - \mu_s) - \mu_s \left( b - a \gamma_m \right) \), we have
    \begin{align*}
        \frac{\partial}{\partial x} f(x) & = \frac{z}{\left((b + \alpha \beta^2) + a x\right)^2} \\
        \frac{\partial^2}{\partial x^2} f(x) & = -\frac{2 a z}{\left((b + \alpha \beta^2) + a x\right)^3}.
    \end{align*}
    The denominators are positive, since they are non-negative and only $ 0 $ if \( \alpha \beta^2 = \gamma_m = b = x = 0 \) (which is the null solution). Therefore, \( f(x) \) is (i) concave increasing if \( z > 0 \), (ii) constant if \( z = 0 \), or (iii) convex decreasing if \( z < 0 \). For $ y = (1 - b) - a (1 - \gamma_m) $, we similarly have
    \begin{align*}
        \frac{\partial}{\partial x} g(x) & = \frac{y}{\left((1 - \gamma_m) - x\right)^2} \\
        \frac{\partial^2}{\partial x^2} g(x) & = \frac{2 y}{\left((1 - \gamma_m) - x\right)^3}.
    \end{align*}
    Since the denominators are positive, \( g(x) \) is (i) convex increasing if \( y > 0 \), (ii) constant if \( y = 0 \), or (ii) concave decreasing if \( y < 0 \). 
    However \( y \geq 0 \):
    \begin{align*}
        y & = (1 - b) - a (1 - \gamma_m) \\
        & = (1 - a) - (b - a \gamma_m) \\
        & \geq 0
    \end{align*}
    since \( a = \tau(\gamma_m) \leq 1 \) and \( b = \mu_{sq}^*(\gamma_m) \geq \tau(\gamma_m) \gamma_m \) as \( \mu_{sq}^*(\gamma_m) \) is concave increasing. We achieve \( y = 0 \) only if \( a = 1, b = \gamma_m = 0 \); if \( y = 0 \), we also have \( z \geq 0 \).
    
    Therefore, there are only two scenarios to consider:
    \begin{enumerate}
        \item \( z \geq 0 \): \( f(x) \) is constant or increasing, \( g(x) \) is constant or increasing \textrightarrow{} \( \gamma^* = \gamma_M \)
        \item \( z < 0 \): \( f(x) \) is convex decreasing, \( g(x) \) is convex increasing. As we will show, \( f(x) g(x) \) is convex \textrightarrow{} \( \gamma^* \in \{\gamma_m, \gamma_M\} \).
    \end{enumerate}
    In the case that \( z < 0 \) (and \( y > 0 \) by the contrapositive), we have
    \begin{align*}
        \argmax_{x \in [0, \gamma_M - \gamma_m]} f(x) g(x) & = \argmax_{x \in [0, \gamma_M - \gamma_m]} \log \left( f(x) g(x) \right) \\
        & = \argmax_{x \in [0, \gamma_M - \gamma_m]} \log \left( \frac{(b - \mu_s \gamma_m) + (a - \mu_s) x}{(b + \alpha \beta^2) + a x} \frac{(1 - b) - ax}{(1 - \gamma_m) - x} \right) \\
        & = \argmax_{x \in [0, \gamma_M - \gamma_m]} \log \left( \frac{c_1 + c_2 x}{c_3 + c_4 x} \frac{c_5 - c_6 x}{c_7 - x} \right)
    \end{align*}   
    We consider the curvature of this:
    \begin{align*}
        \frac{\partial}{\partial x^2} \log \left( \frac{c_1 + c_2 x}{c_3 + c_4 x} \frac{c_5 - c_6 x}{c_7 - x} \right) & = \left[ \frac{c_4^2}{\left( c_3 + c_4 x\right)^2} - \frac{c_2^2}{\left( c_1 + c_2 x\right)^2} \right] + \left[ \frac{1}{\left( c_7 - x\right)^2} - \frac{c_6^2}{\left( c_5 - c_6 x\right)^2} \right] \\
        & > 0 \text{ if } c_1 c_4 > c_2 c_3 \text{ and } c_5 > c_6 c_7
    \end{align*}
    But these conditions are equivalent to \( z < 0 \) and \( g(x) \) increasing:
    \begin{align*}
        \begin{aligned}[t]
            c_1 c_4 - c_2 c_3 & = a (b - \mu_s \gamma_m) - (a - \mu_s) (b + \alpha \beta^2) \\
            & = \mu_s (b - a \gamma_m) - \alpha \beta^2 (a - \mu_s) \\
            & = -z > 0
        \end{aligned}
        \qquad
        \begin{aligned}[t]
            c_5 - c_6 c_7 & = 1 - b - a (1 - \gamma_m) \\
            & = y > 0.
        \end{aligned}
    \end{align*}
    Since we are maximizing a convex function over a bounded domain, the maximizer is one of the domain endpoints, which implies the endpoints of all of the segments (except for possibly \( \gamma_M = 0.5 \)) correspond to categorical solutions. Therefore, the only possible optimal non-categorical solution, where \( \rho > 0 \), occurs when the constraint \( \mu_q \leq 0.5 \) is tight.
\end{proof}

Armed with the optimal form of $ p_s, p_q$, we now derive the solution for the illustrative scenario shown in~\cref{fig:overlapping_sets}, where there is noise in both $ s $ and $ q $. Specifically, we consider when the anomalous and normal sets, respectively $ A $ and $A^c$, might overlap in the data inputs $ s $ and $ q $. \cref{thm:overlapping_sets} shows that, under some mild conditions, the optimal solution always labels the noiseless parts of $ s $ and $ q $ according to their true cluster labels.
The noisy part of $ q $ (i.e., the set $C$) is always labeled as anomalous, but the label for the noisy part of $ s $ (i.e., the set $B$) depends on the setting of $ \beta $. Choosing \( \beta = 0 \) (i.e., \( \hat{F}_0 = \hat{P} \)) prioritizes precision, so the optimal solution does not assign the noisy examples in \( B \) to the anomalous class; choosing \( \beta = \infty \) (i.e., \( \hat{F}_\infty = \hat{R} \)) prioritizes recall, so the optimal solution labels \( B \) as anomalous. This trade-off occurs abruptly at a critical \( \beta_\text{crit} \) that depends on the noise level in both $ s $ and $ q $.

\begin{toappendix}
\begin{lemma}\label{thm:soln_compare}
    Suppose two solutions with \( \mu_{sq}^{(1)}, \mu_{sq}^{(2)} \) and \( D^{(1)}, D^{(2)} \) respectively, where \( D = \frac{\mu_s - \mu_{sq}}{1 - \mu_s} \frac{\mu_q - \mu_{sq}}{1 - \mu_q} \). Assume that \( \mu_{sq}^{(1)} \geq D^{(1)} \) and \( \mu_{sq}^{(2)} \geq D^{(2)} \). Wlog let \( \mu_{sq}^{(1)} \geq \mu_{sq}^{(2)} \). Then, \( \hat{F}_\beta^{(1)} > \hat{F}_\beta^{(2)} \) iff \( z > 0, \beta^2 > \beta_\text{crit}^2 \), where
    \( z  = \left(\mu_{sq}^{(1)} - \mu_{sq}^{(2)}\right) - \left(D^{(1)} - D^{(2)}\right) \) and \( \beta_\text{crit}^2 = \frac{\mu_{sq}^{(2)} D^{(1)} - \mu_{sq}^{(1)} D^{(2)}}{\alpha z} \). Moreover, \( \hat{F}_\beta^{(1)} = \hat{F}_\beta^{(2)} \) iff \( \mu_{sq}^{(1)} = \mu_{sq}^{(2)}, D^{(1)} = D^{(2)} \).
\end{lemma}
\begin{proof}
    Using the alternate definition of \( \hat{F}_\beta \), we see that \( \hat{F}_\beta^{(1)} \geq \hat{F}_\beta^{(2)} \) iff
    \begin{align*}
        (1 + \beta^2) \frac{\mu_{sq}^{(1)} - D^{(1)}}{\mu_{sq}^{(1)} + \alpha \beta^2} & \geq (1 + \beta^2) \frac{\mu_{sq}^{(2)} - D^{(2)}}{\mu_{sq}^{(2)} + \alpha \beta^2}.
    \end{align*}
    This is equivalent to
    \begin{align*}
         \alpha \beta^2 \left(\mu_{sq}^{(1)} - \mu_{sq}^{(2)} + D^{(2)} - D^{(1)} \right) & \geq \mu_{sq}^{(2)} D^{(1)} - \mu_{sq}^{(1)} D^{(2)}.
    \end{align*}
    Let \( z = \mu_{sq}^{(1)} - \mu_{sq}^{(2)} + D^{(2)} - D^{(1)} \), and \( c = \mu_{sq}^{(2)} D^{(1)} - \mu_{sq}^{(1)} D^{(2)} \). We can also write \( c = \left( \mu_{sq}^{(1)} - \mu_{sq}^{(2)} \right) \left(\mu_{sq}^{(1)} - D^{(1)} \right) - \mu_{sq}^{(1)} z \). The first term is always at least $ 0 $ by assumption. Therefore, if \( z \leq 0 \), we must have \( c \geq 0 \), and \( \hat{F}_\beta^{(1)} \) is never strictly better; it is equal iff \( z = 0, \mu_{sq}^{(1)} = \mu_{sq}^{(2)} \). \( \hat{F}_\beta^{(1)} \) is strictly better iff \( z > 0, \alpha \beta^2 > \frac{c}{z} \).    
\end{proof}
\end{toappendix}

\begin{figure}[htbp]
    \centering
    \begin{minipage}{0.4\textwidth}
    \scalebox{0.75}{
        \begin{tikzpicture}
          [every node/.style={align=center,thick}, square/.style={regular polygon,regular polygon sides=4,minimum size=4.25cm}]
          \node (inner1) [draw, square] {\(A^c\)};
          \node (inner2) [draw, square, yshift=2cm, xshift=2cm] {\(A\)};
          \draw [dotted,thick] (-1.5, 0.5) -- (0.5, 0.5);
          \draw [dotted,thick] (0.5, -1.5) -- (0.5, 0.5);
          \draw [dotted,thick] (1.5, 1.5) -- (3.5, 1.5);
          \draw [dotted,thick] (1.5, 1.5) -- (1.5, 3.5);
          \draw [decorate,decoration={brace,mirror,amplitude=6pt}] (-1.48,-1.7) -- (0.48,-1.7) node[midway,yshift=-2em] {\(s(A^c \setminus B)\)};
          \draw [decorate,decoration={brace,mirror,amplitude=6pt}] (0.52,-1.7) -- (1.48,-1.7) node[midway,yshift=-2em] {\(s(B)\)};
          \draw [decorate,decoration={brace,mirror,amplitude=6pt}] (1.52,-1.7) -- (3.48,-1.7) node[midway,yshift=-2em] {\(s(A \setminus B)\)};
          \draw [decorate,decoration={brace,amplitude=6pt}] (-1.7,-1.48) -- (-1.7,0.48) node[midway,xshift=-3em] {\(q(A^c \setminus C)\)};
          \draw [decorate,decoration={brace,amplitude=6pt}] (-1.7,0.52) -- (-1.7,1.48) node[midway,xshift=-3em] {\(q(C)\)};
          \draw [decorate,decoration={brace,amplitude=6pt}] (-1.7,1.52) -- (-1.7,3.48) node[midway,xshift=-3em] {\(q(A \setminus C)\)};
        \end{tikzpicture}
    }        
    \end{minipage}\hspace{1em}
    \begin{minipage}{0.53\textwidth}
    \caption{An illustrative, noisy anomaly scenario which is the setting of \cref{thm:overlapping_sets}. Set $ A $ denotes the anomalous set and $ A^c$ the normal set. We suppose these sets overlap in the data input $ s $ (denoted set $ B$) and input $q$ (denoted set $C$).}
    \label{fig:overlapping_sets}
    \end{minipage}    
\end{figure}

\begin{theoremrep}\label{thm:overlapping_sets}
    Suppose the variable $x$ exists in some probability space $(\Omega, \mathcal F, \mathbb{P})$. As depicted in \cref{fig:overlapping_sets}, denote $A \in \mathcal F$ and $A^c$ its complement, where $ \alpha = \p{A} \leq 0.5 $. Let the sets $ B $ and $ C $ be such that $s(A) \cap s(A^c) = s(B) $ and $ q(A) \cap q(A^c) = q(C) $. Assume that $s$ and $q$ are independent when $x \in A$ (and similarly when $x \in A^c$). Wlog let $ \p{A \setminus B} \geq \p{A \setminus C} $. Then, under mild conditions on the sets $ A, B$, and $C$ (see Appendix), the maximum of $ \hat{F}_\beta $ is achieved when 
    \begin{align*}
        p_s(s(A \setminus B)) & \equiv 1, &
        p_s(s(A^c \setminus B))) & \equiv 0, &
        p_s(s(B))) & \equiv \mathbf{1}\left\{ \beta^2 \geq \beta_\text{crit}^2 \right\}, \\
        p_q(q(A \setminus C)) & \equiv 1, &
        p_q(q(A^c \setminus C)) & \equiv 0, &
        p_q(q(C))) & \equiv 1,
    \end{align*}
    where \( \beta_\text{crit} \) depends on the sets \( A, B \), and \( C \).
\end{theoremrep}
\begin{proof}
    We additionally assume several mild conditions on the sets $A, B$, and $ C$:
    \begin{enumerate}
        \item $ \p{A \cup B}, \p{A \cup C} \leq 0.5 $
        \item $ \p{A \setminus B \setminus C} > 0 $
        \item $ \p{A^c \setminus B \setminus C} \geq \p{A^c \setminus B} \p{A^c \setminus C} $
        \item \( \p{A^c \setminus B \setminus C}\p{A \setminus C} \geq \p{A \setminus B \setminus C} \p{A^c \setminus C} \)
    \end{enumerate}

    Now we define
    \begin{align*}
        \p{A \setminus B \setminus C} & = c_1 \p{A} & \p{A^c \setminus B \setminus C} & = c_4 \p{A^c} \\
        \p{A \setminus B} & = (c_1 + c_2) \p{A} & \p{A^c \setminus B} & = (c_4 + c_5) \p{A^c} \\
        \p{A \setminus C} & = (c_1 + c_3) \p{A} & \p{A^c \setminus C} & = (c_4 + c_6) \p{A^c} \\
        d_1 & = \frac{0.5 - \p{A \cup B}}{\p{A^c \setminus B}} & d_2 & = \frac{0.5 - \p{A \cup C}}{\p{A^c \setminus C}} \\
        d_3 & = \frac{0.5}{\p{A^c \setminus B}} & d_4 & = \frac{0.5}{\p{A^c \setminus C}}.
    \end{align*} 

    At first glance, assigning the sets to labels looks combinatorially difficult, since it is unclear in what order we should assign labels to sets and if any set(s) should be always be labeled the same. We proceed by eliminating most possibilities. By \cref{thm:near_cat} and \cref{thm:cat_tightness}, specifically the form of \( p_q^*(q) \) (depending on \( w(q) = \e_{s|q \in \mathcal{D}_{s|q}}{p_s(s)} \)) and the (near) categorical optimal, we can determine which forms of the solution are possible.
    
    Consider a fixed \( p_s(s) \) assignment with \( p_s(s(A \setminus B)) = v_1, p_s(s(B)) = v_2, p_s(s(A^c \setminus B)) = v_3 \). The question is then in what order do we label \( q(A \setminus C), q(C), q(A^c \setminus C) \). Using the independence of $ s $ and $ q $ within $ A $ and $ A^c $ (i.e., \( \frac{c_1}{c_1 + c_3} = \frac{c_2}{1 - (c_1 + c_3)} \)), we have
    \begin{align*}
        w(q(A \setminus C)) & = \frac{v_1 c_1 + v_2 c_3}{c_1 + c_3} = \frac{v_1 c_2 + v_2 (1 - (c_1 + c_2 + c_3))}{1 - (c_1 + c_3)} \\
        w(q(A^c \setminus C)) & = \frac{v_3 c_4 + v_2 c_6}{c_4 + c_6} = \frac{v_3 c_5 + v_2 (1 - (c_4 + c_5 + c_6))}{1 - (c_4 + c_6)} \\
        w(q(C)) & = \frac{\left( v_1 c_2 + v_2 (1 - c_1 - c_2 - c_3) \right) \p{A} + \left( v_3 c_5 + v_2 (1 - c_4 - c_5 - c_6) \right) \p{A^c}}{(1 - (c_1 + c_3)) \p{A} + (1 - (c_4 + c_6)) \p{A^c}}.
    \end{align*}
    Note that \( w(q(C)) \) is a particular combination of \( w(q(A \setminus C)) \) and \( w(q(A^c \setminus C)) \) such that either \( w(q(A \setminus C)) \geq w(q(C)) \geq w(q(A^c \setminus C)) \) or \( w(q(A \setminus C)) \leq w(q(C)) \leq w(q(A^c \setminus C)) \). Therefore, for any setting of \( v_1, v_2, v_3 \), we will label in either the order \( q(A \setminus C), q(C), q(A^c \setminus C) \) or \( q(A^c \setminus C), q(C), q(A \setminus C) \).
    
    By flipping the role of \( p_s \) and \( p_q \) and applying this logic again, we need only consider solutions in $ s $ and $ q $ that label in this manner. Moreover, $ s $ and $ q $ will label in the same order. Concretely, suppose we labeled $ s $ in the order \( s(A \setminus B), s(B), s(A^c \setminus B) \), with one of the labeling possibilities respecting the constraint \( \mu_s \leq 0.5 \): (i) \( v_1 = 1, v_2 = v_3 = 0 \); (ii) \( v_1 = v_2 = 1, v_3 = 0 \); or (iii) \( v_1 = v_2 = 1, v_3 = d_2 \). Then, we also label $ q $ in the order \( q(A \setminus C), q(C), q(A^c \setminus C) \)) as \( w(q(A \setminus C)) \geq w(q(A^c \setminus C)) \) under each possibility.
    
    As a result, there are only 8 possible optimal solutions, that can be split into 3 groups:
    \begin{itemize}
        \item \( p_s(s(A \setminus B)) = p_q(q(A \setminus C)) = 1, p_s(s(B)) = \rho, p_q(q(C)) = \eta, p_s(s(A^c \setminus B)) = p_q(q(A^c \setminus C)) = 0 \): Four possible solutions with \( \rho^*, \eta^* \in\{0,1\} \).
        \item \( p_s(s(A \cup B)) = p_q(q(A \cup C)) = 1, p_s(s(A^c \setminus B)) = \rho, p_q(q(A^c \setminus C)) = \eta \): Three additional possible solutions with \( (\rho^*, \eta^*) \in \{(0, d_2), (d_1, 0), (d_1, d_2)\} \).
        \item \( p_s(s(A \cup B)) = p_q(q(A \cup C)) = 0, p_s(s(A^c \setminus B)) = \rho, p_q(q(A^c \setminus C)) = \eta \): One only possible solution \( \rho^* = d_3, \eta^* = d_4 \).
    \end{itemize}

    Let's consider the first group, with $ p_s(s(A \setminus B)) = p_q(q(A \setminus C)) = 1, p_s(s(A^c \cup B)) = \rho, p_q(q(A^c \cup C)) = \eta, p_s(s(A^c \setminus B)) = p_q(q(A^c \setminus C)) = 0 $. We have
    \begin{align*}
        \mu_{sq} & = \left[ c_1 + \rho c_3 + \eta c_2 + \rho \eta (1 - c_1 - c_2 - c_3) \right] \p{A} + \rho \eta (1 - c_4 - c_5 - c_6) \p{A^c} \\
        \mu_s & = \left[ (c_1 + c_2) + \rho (1 - c_1 - c_2) \right] \p{A} + \rho (1 - c_4 - c_5) \p{A^c} \\
        \mu_q & = \left[ (c_1 + c_3) + \eta (1 - c_1 - c_3) \right] \p{A} + \eta (1 - c_4 - c_6) \p{A^c} \\
        \frac{\mu_s - \mu_{sq}}{1 - \mu_s} & = \frac{(1 - \eta) \left[ c_2 + \rho (1 - c_1 - c_2 - c_3) \right] \p{A} + \rho \left[ (1 - \eta) (1 - c_4 - c_5) + \eta c_6 \right] \p{A^c}}{(1 - \rho) (1 - c_1 - c_2) \p{A} + \left[1 - \rho (1 - c_4 - c_5) \right] \p{A^c}} \\
        \frac{\mu_q - \mu_{sq}}{1 - \mu_q} & = \frac{(1 - \rho) \left[ c_3 + \eta (1 - c_1 - c_2 - c_3) \right] \p{A} + \eta \left[ (1 - \rho) (1 - c_4 - c_6) + \rho c_5 \right] \p{A^c}}{(1 - \eta) (1 - c_1 - c_3) \p{A} + \left[1 - \eta (1 - c_4 - c_6) \right] \p{A^c}}
    \end{align*}
    Note that due to the assumption \( \p{A \setminus B} \geq \p{A \setminus C} \), we have \( c_2 \geq c_3 \geq 0 \). To analyze the possible solutions (\( \rho^*, \eta^* \in\{0,1\} \)), we consider the alternate definition for \( \hat{F}_\beta \)
    \begin{align*}
        \hat{F}_\beta & = (1 + \beta^2) \frac{\mu_{sq} - D}{\mu_{sq} + \alpha \beta^2} \\
        D & = \frac{\mu_s - \mu_{sq}}{1 - \mu_q} \frac{\mu_q - \mu_{sq}}{1 - \mu_s}
    \end{align*}
    under the four options:
    \begin{itemize}
        \item \( \rho^* = \eta^* = 0 \): Since \( \mu_s, \mu_q \geq \mu_{sq} \), we have \( D \geq 0 \) and
        \begin{align*}
            \hat{F}_\beta & = (1 + \beta^2) \frac{c_1 \p{A} - D}{c_1 \p{A} + \alpha \beta^2}
        \end{align*}
        \item \( \rho^* = 1, \eta^* = 0 \): Since \( \mu_{sq} = \mu_q \), we have \( D = 0 \) and 
        \begin{align*}
            \hat{F}_\beta & = (1 + \beta^2) \frac{(c_1 + c_3) \p{A}}{(c_1 + c_3) \p{A} + \alpha \beta^2}
        \end{align*}
        \item \( \rho^* = 0, \eta^* = 1 \): Since \( \mu_{sq} = \mu_s \), we have \( D = 0 \) and 
        \begin{align*}
            \hat{F}_\beta & = (1 + \beta^2) \frac{(c_1 + c_2) \p{A}}{(c_1 + c_2) \p{A} + \alpha \beta^2}
        \end{align*}
        \item \( \rho^* = \eta^* = 1 \)
        \begin{align*}
            \hat{F}_\beta & = (1 + \beta^2) \frac{\p{A} + (1 - c_4 - c_5 - c_6) \p{A^c} - D}{\p{A} + (1 - c_4 - c_5 - c_6) \p{A^c} + \alpha \beta^2} \\
            D & = \frac{c_5}{c_4 + c_5} \frac{c_6}{c_4 + c_6}
        \end{align*}
    \end{itemize}
    Since \( D(\rho^*=0,\eta^*=0) \geq 0 \), we have
    \begin{align*}
        \hat{F}_\beta (\rho^*=1, \eta^*=0) \geq \hat{F}_\beta (\rho^*=0, \eta^*=0)
    \end{align*}
    where the inequality is equality iff \( c_3 = 0 \).
    Furthermore, since \( c_2 \geq c_3 \geq 0 \), we have
    \begin{align*}
        \hat{F}_\beta (\rho^*=0, \eta^*=1) \geq \hat{F}_\beta (\rho^*=1, \eta^*=0)
    \end{align*}
    where the inequality is equality iff \( \beta = 0 \) or \( c_2 = c_3 \).
    The relationship between \( \hat{F}_\beta (\rho^*=0, \eta^*=1) \) and \( \hat{F}_\beta (\rho^*=1, \eta^*=1) \) depends on the various probabilities and also the setting of \( \beta \). Specifically, by \cref{thm:soln_compare}, we have
    \begin{align*}
        \hat{F}_\beta (\rho^*=1, \eta^*=1) > \hat{F}_\beta (\rho^*=0, \eta^*=1)
    \end{align*}
    iff \( z > 0 \) and \( \beta^2 > \beta_\text{crit}^2 \) where
    \begin{align*}
        z & = \left( \mu_{sq}(\rho^*=1, \eta^*=1) - \mu_{sq}(\rho^*=0, \eta^*=1) \right) - \left(D(\rho^*=1, \eta^*=1) - D((\rho^*=0, \eta^*=1) \right) \\
        & = \left((1 - c_1 - c_2) \p{A} + (1 - c_4 - c_5 - c_6) \p{A^c} \right) - \frac{c_5}{c_4 + c_5} \frac{c_6}{c_4 + c_6} \\
        & = \p{A \cap B} + \p{A^c \cap B \cap C} - \frac{\p{(A^c \setminus B) \cap C}}{\p{A^c \setminus B}} \frac{\p{(A^c \setminus C) \cap B}}{\p{A^c \setminus C}} \\
        \beta_\text{crit}^2 & = \frac{\mu_{sq}(\rho^*=0, \eta^*=1) D(\rho^*=1, \eta^*=1) - \mu_{sq}(\rho^*=1, \eta^*=1) D(\rho^*=0, \eta^*=1)}{\alpha z} \\
        & = \frac{(c_1 + c_2) \p{A}}{\alpha z} \frac{c_5}{c_4 + c_5} \frac{c_6}{c_4 + c_6} \\
        & = \frac{\p{A \setminus B}}{\p{A}} \frac{\p{(A^c \setminus B) \cap C}}{\p{A^c \setminus B}} \frac{\p{(A^c \setminus C) \cap B}}{\p{A^c \setminus C}} \frac{1}{z}
    \end{align*}
    We have equality iff \( c_5 c_6 = 0 \) (i.e., \( B \) or \( C \) is the empty set); but by assumption (\( \p{A \setminus B} \geq \p{A \setminus C} \)), this must be satisfied first by \( B = \emptyset \). In this case, we don't care about the setting of \( \rho \). So whenever we care to label \( s(B) \), the condition on \( z \) is always met under the assumption \( \p{A^c \setminus B \setminus C} \p{A \setminus C} \geq \p{A \setminus B \setminus C} \p{A^c \setminus C} \):
    \begin{align*}
        z & = \p{A \cap B} + \p{A^c \cap B \cap C}  - \frac{\p{(A^c \setminus B) \cap C}}{\p{A^c \setminus B}} \frac{\p{(A^c \setminus C) \cap B}}{\p{A^c \setminus C}} \\
        & = \p{A \cap B} + \p{A^c \cap B \cap C}  - \frac{\p{(A^c \setminus B) \cap C}}{\p{A^c \setminus B}} \frac{\p{(A^c \setminus C) \cap B}}{\p{A^c \setminus C}} \\
        & = \p{A \cap B} - \left( 1 - \p{A^c} \right) \frac{\p{(A^c \setminus B) \cap C}}{\p{A^c \setminus B}} \frac{\p{(A^c \setminus C) \cap B}}{\p{A^c \setminus C}} \\
        & = \p{A \cap B} - \p{A} \frac{\p{(A^c \setminus B) \cap C}}{\p{A^c \setminus B}} \frac{\p{(A^c \setminus C) \cap B}}{\p{A^c \setminus C}} \\
        & = \p{A} \left[ \left(1 - \frac{\p{A \setminus B \setminus C}}{\p{A \setminus C}} \right) - \left(1 - \frac{\p{A^c \setminus B \setminus C}}{\p{A^c \setminus B}} \right) \left( 1 - \frac{\p{A^c \setminus B \setminus C}}{\p{A^c \setminus C}} \right) \right] \\
        & = \p{A} \left[ \frac{\p{A^c \setminus B \setminus C}}{\p{A^c \setminus C}} - \frac{\p{A \setminus B \setminus C}}{\p{A \setminus C}} + \left(1 - \frac{\p{A^c \setminus B \setminus C}}{\p{A^c \setminus C}} \right) \frac{\p{A^c \setminus B \setminus C}}{\p{A^c \setminus B}} \right] \\
        & > 0
    \end{align*}
    where we used the equalities
    \begin{align*}
        \frac{\p{A^c \cap B \cap C}}{\p{A^c \cap B}} & = \frac{\p{(A^c \setminus B) \cap C}}{\p{A^c \setminus B}} \\
        \frac{\p{A^c \cap B}}{\p{A^c}} & = \frac{\p{(A^c \setminus C) \cap B}}{\p{A^c \setminus C}} \\
        \frac{\p{A \cap B}}{\p{A}} & = \frac{\p{(A \setminus C} \cap B}{\p{A \setminus C}}
    \end{align*}
    since $ s $ and $ q $ are independent within $ A $ and $ A^c $.


    Now, let's consider the second group, with $ p_s(s(A \cup B)) = p_q(q(A \cup C)) = 1, p_s(s(A^c \setminus B)) = \rho, p_q(q(A^c \setminus C)) = \eta $. We can write
    \begin{align*}
        \mu_{sq} & = \p{A} + \left[ (1 - c_4 - c_5 - c_6) + \rho c_5 + \eta c_6 + \rho \eta c_4 \right] \p{A^c} \\
        \mu_s & = \p{A} + \left[ (1 - c_4 - c_5) + \rho (c_4 + c_5) \right] \p{A^c} \\
        \mu_q & = \p{A} + \left[ (1 - c_4 - c_6) + \eta (c_4 + c_6) \right] \p{A^c} \\
        D & = \frac{c_6 + \rho c_4}{c_4 + c_6} \frac{c_5 + \eta c_4}{c_4 + c_5}
    \end{align*}
    We again analyze the possible solutions (\( (\rho^*, \eta^*) \in \{(0, d_2), (d_1, 0), (d_1, d_2)\} \)), and compare to \( (\rho^* = \eta^*=0) \) for improvement:
    \begin{itemize}
        \item \( \rho^* = 0, \eta^* = d_2 \): By \cref{thm:soln_compare} (the condition on \( z \)), this solution cannot be an improvement if
        \begin{align*}
            \mu_{sq}^*(\rho^* = 0, \eta^* = d_2) - \mu_{sq}^*(\rho^* = 0, \eta^* = 0) & \leq D(\rho^* = 0, \eta^* = d_2) - D(\rho^* = 0, \eta^* = 0) \\
            d_2 c_6 \p{A^c} & \leq d_2 c_6 \frac{1}{c_4 + c_6} \frac{c_4}{c_4 + c_5} \\
            \p{A^c} & \leq \frac{1}{c_4 + c_6} \frac{c_4}{c_4 + c_5},
        \end{align*}
        which is equivalently \( \p{A^c \setminus B \setminus C} \geq \p{A^c \setminus B} \p{A^c \setminus C} \). This holds by assumption. The solutions are equal iff \( c_6 = 0 \).
        \item \( \rho^* = d_1, \eta^* = 0 \): By symmetry, this is not an improvement. The solutions are equal iff \( c_5 = 0 \).
        \item \( \rho^* = d_1, \eta^* = d_2 \): In fact, the same condition guarantees that this solution cannot be an improvement either, as we again have
        \begin{align*}
            \mu_{sq}^*(\rho^* = d_1, \eta^* = d_2) - \mu_{sq}^*(\rho^* = 0, \eta^* = 0) & \leq D(\rho^* = 0, \eta^* = d_2) - D(\rho^* = 0, \eta^* = 0) \\
            \left[ d_1 c_5 + d_2 c_6 + d_1 d_2 c_4 \right] \p{A^c} & \leq \left[d_1 c_5 + d_2 c_6 + d_1 d_2 c_4\right] \frac{c_4}{(c_4 + c_6) (c_4 + c_5)} \\
            \p{A^c} & \leq c_4 \frac{c_4}{(c_4 + c_6) (c_4 + c_5)}
        \end{align*}
    \end{itemize}
    None of these solutions are improvements (or even equal except under corner cases) and therefore can be ignored.

    Finally, let's consider the third group, with just the one possible solution: \( p_s(s(A \cup B)) = p_q(q(A \cup C)) = 0, p_s(s(A^c \setminus B)) = d_3, p_q(q(A^c \setminus C)) = d_4 \). We will compare this to $ p_s(s(A \cup B)) = p_q(q(A \cup C)) = 1, p_s(s(A^c \setminus B)) = d_1, p_q(q(A^c \setminus C)) = d_2 $. We can in fact we can show that the \( \mu_{sq} \) for both solutions are equal: \( d_3 d_4 c_4 \p{A^c} = \p{A} + \left[ (1 - c_4 - c_5 - c_6) + d_1 c_5 + d_2 c_6 + d_1 d_2 c_4 \right] \p{A^c} \) since \( 1 - d_1 = d_3, 1 - d_2 = d_4 \). Moreover, since both solutions achieve \( \mu_s = \mu_q = 0.5 \), they are equal for all \( \beta \). But since the solution we are comparing against is never optimal, this is never optimal either!
\end{proof}

Lastly, it is worth noting that we can interpret our method, in the continuous case, as performing both feature representation and classification, simultaneously and in an unsupervised manner. We expand upon this further in the Appendix.

\begin{toappendix}
\subsection{Simultaneous feature representation and classification}
In the case where both models \( A_{\theta_s}, A_{\theta_q} \) are parameterized as DNNs, we note that our method performs both feature representation and classification, simultaneously and in an unsupervised manner. Suppose the final components of each DNN are a fully-connected (FC) layer and a sigmoid activation. The final layer and sigmoid then amount to a logistic classifier on the latent representations (i.e., representation fed to the final FC layer). The backbones of the networks are then tasked with learning how to generate linearly-separable representations of the input data. \cref{thm:grads} shows the gradients with respect to the logistic classifier parameters $ w_s, b_s $ and the latent representations $ z_s(s) $, and illustrates several important qualitative behaviors. First, the gradients depend \textit{most} on the uncertain examples \( p_{s}(s) \) and ignore examples with high confidence, as \( \gamma_s(s) = 0 \) for \( p_s(s) \in \{0,1\} \), meaning the examples on the “margin” are most important. Second, the gradients also depend on the pseudo-target \( \hat{y}_q(q) \), which has largest magnitude for the most certain examples \( p_{q}(q) \). Taken together, the \( \hat{F}_\beta \) metric encourages each individual model to adopt the other's prediction when it is uncertain but the other model is confident.
\begin{theoremrep}\label{thm:grads}
    Suppose \( p_s, p_q \) are parameterized as DNNs, whose final components of each DNN are a fully-connected layer and a sigmoid activation. Let the latent representation fed into the final layer be \( z_s(s), z_q(q) \in \mathbb{R}^{p} \) for the two networks respectively. Define \( p_s(s) = \sigma(w_s^T z_s(s) + b_s) \) and \( p_q(q) = \sigma(w_q^T z_q(q) + b_q) \). Then,
    \begin{align*}
        \nabla_{w_s} \hat{F}_\beta & \equiv \e_{\mathcal{D}}{\hat{y}_q(q) \gamma_s(s) z_s(s)}, \\
        \nabla_{b_s} \hat{F}_\beta & \equiv \e_{\mathcal{D}}{\hat{y}_q(q) \gamma_s(s)}, \\
        \nabla_{z_s(s)} \hat{F}_\beta & \equiv \e_{\mathcal{D}_{q|s}}{\hat{y}_q(q)} \gamma_s(s) w_s,
    \end{align*}
    where \( \hat{y}_p(p) = c_1 p_q(q) - c_2 \) and \( \gamma_s(s) = p_s(s) \left(1 - p_s(s)\right) \). The gradients w.r.t.~the $ q $ network are analogous. The constants \( c_1, c_2 \) depend on the current parameters \( \theta_s, \theta_q \) but not individual instances \( (s,q) \in \mathcal{D} \). If the networks are better than random guessers, then \( c_1, c_2 \geq 0 \).
\end{theoremrep}
\begin{proof}
    Recall that for \( y = \sigma(a^T x + b) \), we have \( \nabla_a y = y (1-y) x \) and \( \nabla_b y = y (1-y) \). Let us consider the gradients w.r.t.~the parameters of the linear classifier:
    \begin{align*}
        \nabla_{w_s} \hat{F}_\beta & = \nabla_{w_s} \left( (1 + \beta^2) \frac{\mu_{sq} - \mu_s \mu_q}{\mu_{sq} + \alpha \beta^2} \frac{1 - \mu_{sq}}{(1 - \mu_s)(1 - \mu_q)} \right) \\
        & = \hat{F}_\beta \left[ \frac{\nabla_{w_s} (1 - \mu_{sq})}{1 - \mu_{sq}} + \frac{\nabla_{w_s} (\mu_{sq} - \mu_s \mu_q)}{\mu_{sq} - \mu_s \mu_q} \right] \\
        & - \hat{F}_\beta \left[\frac{\nabla_{w_s} (\mu_{sq} + \alpha \beta^2)}{\mu_{sq} + \alpha \beta^2} + \frac{\nabla_{w_s} (1 - \mu_s)}{1 - \mu_s} \right] \\
        & = \hat{F}_\beta \left[ \frac{1}{\mu_{sq} - \mu_s \mu_q} - \frac{1}{1 - \mu_{sq}} - \frac{1}{\mu_{sq} + \alpha \beta^2} \right] \e_{\mathcal{D}}{p_q(q) \nabla_{w_s} p_s(s) } \\
        & - \hat{F}_\beta \left[ \frac{\mu_q}{\mu_{sq} - \mu_s \mu_q} - \frac{1}{1 - \mu_s} \right] \e_{\mathcal{D}}{\nabla_{w_s} p_s(s)} \\
        & = \e_{\mathcal{D}}{(c_1 p_q(q) - c_2) \nabla_{w_s} p_s(s)} \\
        & = \e_{\mathcal{D}}{(c_1 p_q(q) - c_2) p_s(s) (1 - p_s(s)) \nabla_{w_s} (w_s^T z_s(s) + b_s)} \\
        & \equiv \e_{\mathcal{D}}{\hat{y}_q(q) \gamma_s(s) z_s(s)}
    \end{align*}
    where we use the dominating convergence theorem to exchange the derivative and expectation (note that the expectations are finite and \( \mu_{sq}, \mu_s \) are differentiable for all choices of weights; moreover the derivatives are bounded). We have defined
    \begin{align*}
        \gamma_s(s) & = p_s(s) (1 - p_s(s)) \\
        \hat{y}_p(p) & = c_1 p_q(q) - c_2 \\
        c_1 & = \hat{F}_\beta \left[ \frac{1}{\mu_{sq} - \mu_s \mu_q} - \frac{1}{1 - \mu_{sq}} - \frac{1}{\mu_{sq} + \alpha \beta^2} \right] \\
        c_2 & = \hat{F}_\beta \left[ \frac{\mu_q}{\mu_{sq} - \mu_s \mu_q} - \frac{1}{1 - \mu_s} \right] \\
        & = \hat{F}_\beta \frac{\mu_q - \mu_{sq}}{(\mu_{sq} - \mu_s \mu_q) (1 - \mu_s)}
    \end{align*}
    The gradients w.r.t.~the latent representation \( z_s(s) \) under the \( \hat{F}_\beta \) loss is
    \begin{align*}
        \nabla_{z_s(s)} \hat{F}_\beta & = \e_{\mathcal{D}_{q|s}}{(c_1 p_q(q) - c_2) p_s(s) (1 - p_s(s)) w_s} \\
        & \equiv \e_{\mathcal{D}_{q|s}}{\hat{y}_q(q)} \gamma_s(s) w_s
    \end{align*}
    Lastly, we can show that \( c_1, c_2 \geq 0 \) under the assumption that if the networks are better than random guessers (i.e., \( \mu_{sq} \geq \mu_s \mu_q \) ). We have that \( c_1 \geq 0 \):
    \begin{align*}
        c_1 & = \hat{F}_\beta \left[ \frac{1}{\mu_{sq} - \mu_s \mu_q} - \frac{1}{1 - \mu_{sq}} - \frac{1}{\mu_{sq} + \alpha \beta^2} \right] \\
        & = \hat{F}_\beta  \left[ \frac{-(\mu_{sq} - \mu_s \mu_q)(1 - \mu_{sq}) + (\mu_{sq} + \alpha \beta^2) \left(1 - 2 \mu_{sq} + \mu_s \mu_q \right)}{(\mu_{sq} - \mu_s \mu_q) (1 - \mu_{sq}) (\mu_{sq} + \alpha \beta^2)} \right] \\
        & = \hat{F}_\beta  \left[ \frac{\left(\mu_s \mu_q - \mu_{sq}^2\right) + \alpha \beta^2 \left(1 - 2 \mu_{sq} + \mu_s \mu_q \right)}{(\mu_{sq} - \mu_s \mu_q) (1 - \mu_{sq}) (\mu_{sq} + \alpha \beta^2)} \right] \\
        & \geq 0
    \end{align*}
    where \( \mu_s \mu_q \geq \mu_{sq}^2 \) by Cauchy-Schwartz (shown in more detail in the proof of~\cref{thm:near_cat}) and since \( \mu_{sq} \leq 0.5 \). Also, since \( \mu_{sq} \leq \mu_q \) by definition, we have \( c_2 \geq 0 \).
\end{proof}
\end{toappendix}

\section{Related Work}

\subsection{Unsupervised anomaly detection}

In contrast to our coincidence-based approach, existing unsupervised methods fall into three categories: probability-, density-, and compression-based. The most popular probability- and density-based methods---such as GMM~\cite{reynolds2009gaussian}, IF~\cite{Liu08}, OCSVM~\cite{Scholkopf01}, LOF~\cite{Breunig00}, and KDE~\cite{Parzen62}---can struggle with high-dimensional inputs, particularly where the anomalies are both present in the training set and densely clustered together. More recent works leverage DNNs to learn deep feature representations of the inputs. Most common is compression-based methods using autoencoder-type architectures and reconstruction-based anomaly scores~\cite{Chen17, su2019robust, Schlegl17, challu2022deep, zenati2018adversarially, yang2020regularized, schlegl2019f, liu2019generative, zhou2021vae}. Other works combine density and compression approaches by measuring distances to cluster centers in low-dimensional spaces~\cite{Ruff18, shen2020timeseries}. However, these approaches still assume (i) the training data is completely “normal” and/or (ii) the anomalies exist in low-density regions. Our method does not make these assumptions; in fact, we explicitly assume our data is polluted by anomalies, which are quite possibly densely clustered (for example, repeated failures might have extremely similar signatures). Moreover, we expect this setting to be quite prevalent in many practical applications.


\subsection{Common representation learning}
For classification problems in the absence of labels, prior works have proposed unsupervised methods for learning feature representations of different views (or modalities) of the data. The key concept in these approaches is to maximize the alignment between the feature representations, such as maximizing the correlation, covariance, or “semantic similarity” between the learned feature representations. Though not designed for anomaly detection, these methods share similarities with CoAD.

The correlation approach, proposed in Canonical Correlation Analysis (CCA)~\cite{Hotelling_1936}, has many different variants and applications~\cite{Akaho2006,Hardoon2004,Wang2015,Chandar_2016}. Deep Canonical Correlation Analysis (DCCA)~\cite{Andrew2013} proposes (in the notation of our paper) the objective \( \max_{\theta_s, \theta_q} \corr\left(A_{\theta_s}, A_{\theta_q} \right) \) where the models are parameterized as DNNs. 
This correlation objective is qualitatively similar to the special case \( \hat{F}_0 = \hat{P} \) of our method, in that they both promote maximum precision and are agnostic to the number of anomalies found. This behavior of \( \hat{P} \) (and correlation) is why the control over \( \beta \) (i.e., between \( \hat{P}\) and \( \hat{R} \)) is desirable.

Maximum covariance analysis (MCA) (or SVD analysis) ~\cite{Storch1999,Bretherton1992,Wallace1992} instead maximizes the covariance between linear projections of the two views. Deep Maximum Covariance Analysis (DMCA)~\cite{Luo2018} proposes (in the notation of our paper) the objective \( \max_{\theta_s, \theta_q} \cov\left(A_{\theta_s}, A_{\theta_q}\right) \). This covariance objective is qualitatively similar to the special case \( \hat{F}_\infty = \hat{R} \), in that they both promote maximum recall. However, in our setting where $ s $ and $ q $ are independent given the true label (see~\cref{thm:disagree_rate}), covariance can only be a worse underestimate of the recall (i.e., number of anomalies) than our metric \( \hat{R} \). Specifically, we have $
    R \geq \hat{R} = \frac{1}{\alpha} \cov\left(p_s, p_q\right) \frac{1 - \mu_{sq}}{(1 - \mu_s) (1 - \mu_q)} \geq 
\frac{1}{\alpha} \cov\left(p_s, p_q\right)$, where the second inequality follows from \( \mu_{sq} \leq \mu_s, \mu_q \).
We show this for a synthetic example in~\cref{sec:thresh_results}.

Lastly, maximizing the “semantic similarity” between feature representations is commonly done by maximizing the mutual information~\cite{tian2020contrastive,bachman2019learning} or with a contrastive loss~\cite{Chen2020,henaff2020data,shaham2018_learning}.


\section{Experiments}


We assess our method on several test cases, in both the categorical and continuous settings. In the continuous setting, we train our DNNs using the PyTorch framework~\cite{Paszke19}, Adam optimizer~\cite{Kingma14}, mini-batches, and a sigmoid-based regularizer to enforce the constraint $ \mu_s, \mu_q \leq 0.5 $. We also use our unsupervised metric $ \hat{F}_\beta $ as both an early stopping and hyperparameter selection criterion. Full details on the training settings and architectures can be found in the Appendix.

\begin{toappendix}
    In the continuous setting, where the models are represented by DNNs, we need to enforce the constraints \( \mu_s, \mu_ q\leq 0.5 \). In order to use gradient-based optimizers, we impose these constraints using a “wall” regularization term \( \mathcal{L}_\text{wall} = \mu_s \sigma\left(t (\mu_s - 0.5)\right) + \mu_q \sigma\left(t (\mu_q - 0.5)\right) \) where \( t \) is the wall's temperature (set to $50$ for all experiments) and \( \sigma(\cdot) \) is the sigmoid function. Also, to prevent the network from quickly railing to $ \{0,1\} $ (and thereby suffer from vanishing gradients), we impose a norm penalty on the magnitude of the output logits (i.e., outputs prior to the final sigmoid converting to probability): \( \mathcal{L}_\text{mag} = \e{\text{logit}(p_s)^2 + \text{logit}(p_q)^2} \). In total, we minimize the loss
    \begin{align*}
        \mathcal{L} = - \hat{F}_\beta + \lambda_\text{wall} \mathcal{L}_\text{wall} + \lambda_\text{mag} \mathcal{L}_\text{mag}
    \end{align*}
    where \( \lambda_\text{wall}, \lambda_\text{mag} \) are the regularization strengths.
\end{toappendix}

\subsection{Synthetic outliers}\label{sec:thresh_results}
We start by illustrating the categorical case. We create a synthetic dataset of 20k normal points sampled from $|\mathcal{N}(0,1)|$ for both $s$ and $q$. We then introduce anomalies by sampling from $1+|\mathcal{N}(0,1.5)|$ for 5\% of the data points. Anomalies always occur simultaneously in both inputs. We then use a simple threshold as the anomaly detection model, applied independently to each data input, and identify as anomalous any point that exceeds the threshold simultaneously in both $s$ and $q$. This setup corresponds to pre-specified models \( A_{\theta_s}, A_{\theta_q} \) where the parameters are just single thresholds on the model outputs. \cref{fig:synth_data} illustrates a small example.

\begin{figure}[htbp]
    \centering
    \subfigure[{Comparison of $P$-$R$/$\hat{P}$-$\hat{R}$ curves.}]{%
        \includegraphics[width=0.42\linewidth]{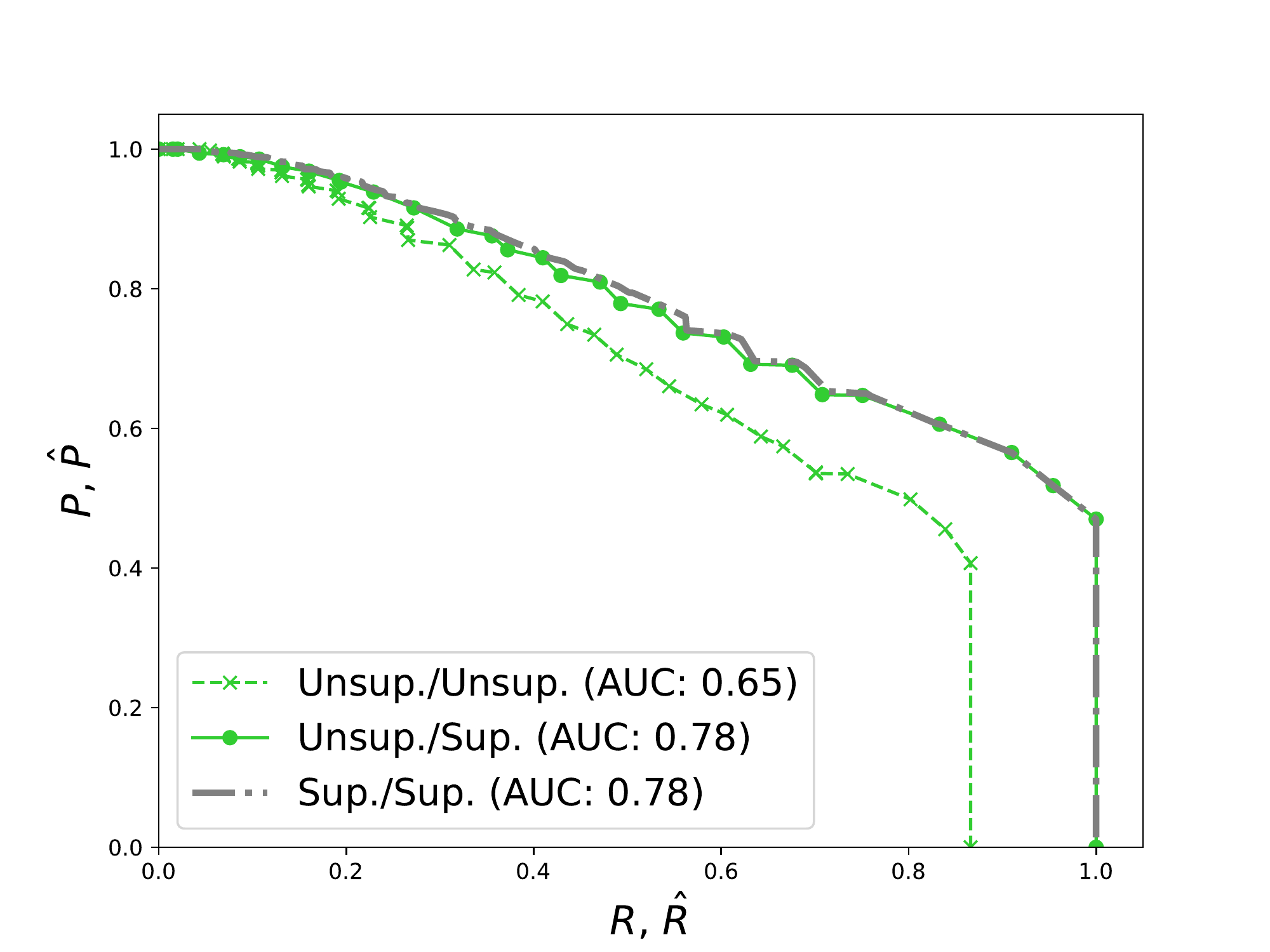}%
        \label{fig:synth_auc}%
    }\hspace{1em}
    \subfigure[{Comparison of $ F_\beta $/$\hat{F}_\beta$ values, for two different disagreement definitions.}]{%
        \includegraphics[width=0.42\linewidth]{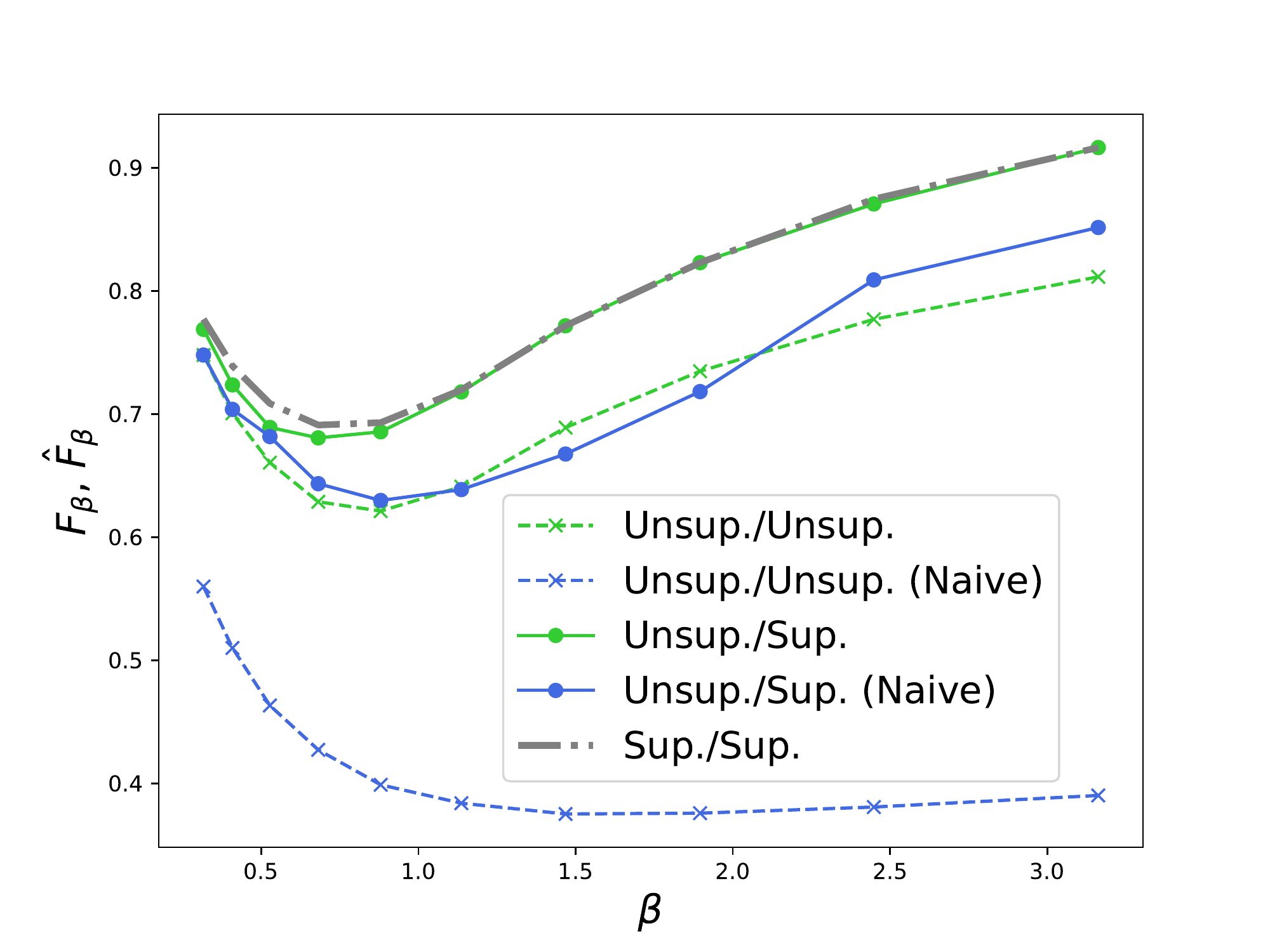}%
        \label{fig:synth_beta_scan}%
    }%
    \caption{Analysis of our unsupervised method on a synthetic dataset, comparing fully supervised, unsupervised train but supervised test, and fully unsupervised cases. Each curve is labeled by its train/test type, where “Unsup.” is unsupervised and “Sup.” is supervised.}
    \label{fig:synth_plots}
\end{figure}

To evaluate our approach, we will compare three different precision-recall curves. The first curve is the unsupervised \( \hat{P} \)-\( \hat{R} \) curve, where we select the thresholds and evaluate without the available ground truth labels (i.e., unsupervised training and testing). The second curve uses the same $ s $ and $ q $ thresholds as the first curve but evaluates against the true labels (i.e., unsupervised training but supervised testing). The last curve is the supervised $P$-$R$ curve, where we select the optimal $s$ and $q$ thresholds and evaluate them using the ground truth labels. \cref{fig:synth_auc} shows these three precision-recall curves, where we plot \( \hat{R}, \hat{P} \) for the unsupervised case. When evaluated on the true labels, our unsupervised method almost exactly matches the supervised precision-recall curve. Also, since the assumptions of~\cref{thm:disagree_rate} are met for this synthetic case, \( \hat{R}, \hat{P} \) are underestimates of recall $R$ and precision $P$, respectively, for any choice of thresholds.


We can also analyze \( \hat{F}_\beta \) and the thresholds that maximize it, for a range of \( \beta \). As above, we have the unsupervised \( \hat{F}_\beta\) curve, the \( F_\beta \) curve from evaluating the unsupervised thresholds with the true labels, and the supervised \( F_\beta \) curve. To illustrate our method for estimating the fraction of false positives, we compare the definitions given in~\cref{thm:disagree_rate,thm:naive_rate}. We denote the latter definition, \( D_\text{naive} \), as the “naive” rate. \cref{fig:synth_beta_scan} shows the different curves, where we plot \( \hat{F}_\beta \) for the unsupervised cases. While both estimates of the false positive fraction lead to underestimates of \( \hat{F}_\beta \), our definition $ \D $ gives a significantly better estimate than $ D_\text{naive} $, and the corresponding $s$ and $q$ thresholds achieve near supervised-level performance when assessed with the true labels.

%

\subsection{MNIST}\label{sec:mnist}
As a second illustration, we construct an image-based anomaly detection task using MNIST. Each example consists of a pair of images, one given as input to each network. Normal examples consist of a pair of images of the digit $0$.  Anomalous examples are pairs of images drawn from the digits $1, 2$, or $3$, with the same digit given to each network (i.e., there are 3 different anomaly types). We then alter the difficulty by imposing a noisy observation model, where with some probability the image fed to each network is replaced by another digit, such that each anomaly class has different amounts of noise ($1$ having the least noise and $3$ having the most noise). The full observation model is detailed in the Appendix.

\begin{table}[htbp]
    \begin{minipage}{.48\columnwidth}
    \begin{tabular}{cccc}
        \toprule Digit &  $\beta=0.05$ & $\beta=1$ & $\beta=\infty$  \\ \midrule
        0 & $0\%$ & $0.5\%$ & $0.5\%$ \\
        1 & $96.3\%$ & $99.1\%$ & $100\%$ \\
        2 & $0.2\%$ & $96.9\%$ & $99.2\%$ \\
        3 & $0.1\%$ & $1.2\%$ & $99.6\%$ \\ \bottomrule
    \end{tabular}\\[1em]
    \end{minipage}
    \begin{minipage}{.51\columnwidth}
    \caption{Fraction of MNIST digits labeled as anomalous under a noisy observation model. A digit is labeled anomalous if the product of the two network outputs is greater than $ 0.5 $. Results are shown for several different values of $\beta$ and rounded to the nearest tenth of a percent.}
    \label{tab:mnist}
    \end{minipage}
\end{table}

\cref{tab:mnist} shows the results for different values of $ \beta $. In particular, we illustrate that different choices of $ \beta $ result in classifying different sets of anomalies: small $ \beta $ only identifies the least noisy anomalies, and large $ \beta $ identifies all 3 anomaly classes. The full violin plots are shown in the Appendix. Also, despite being trained without labels, our method learned to separate the latent representations of the digits, created by the models $ A_{\theta_s} $ and $ A_{\theta_q} $. We visualize these latent spaces in the Appendix.

\begin{toappendix}
    \subsection{MNIST}

    For the MNIST example, we use a noisy observation model to ensure that each anomaly class (digits $1,2$ and $3$) has a different amount of noise. In particular, we suppose that each anomaly digit $i$ has frequency $ w_i $ and a “blur” probability $ b_i $. Then, for each data input, we replace the anomalous digit with the normal digit (i.e., $0$) with probability $ b_i $; we also symmetrically replace the normal digit with the anomalous digit $ i $ with probability $ b_i $. This creates a data set with mixed image pairs. For our experiments, we used the setting $ w_0 = 0.85, w_1 = w_2 = w_3 = 0.05 $ and $ b_1 = 0, b_2 = 0.05, b_3 = 0.2 $. It is worth noting that our experiment is not specific to this particular observation model: other ways of generating mixed image pairs would also work.

    To better understand the behavior of our metric in this setting, we plot $ \hat{F}_\beta $ under the simplification that all digits of the same class are identical. This allows us to derive a closed form for $ \hat{F}_\beta $ as a function of the $ w_i, b_i, \beta $, and the labeling choice $ p_s(\text{digit } i) = p_q(\text{digit } i) = y_i $. We show the $3$ obvious labeling choices in~\cref{fig:mnist_toy}. Each labeling choice is optimal for a range of $ \beta $, and the higher choices of $ \beta $ (i.e., higher recall preference) lead to labeling the noisy classes as anomalous.
    \begin{figure}[htbp]
        \centering
        \includegraphics[width=0.6\linewidth]{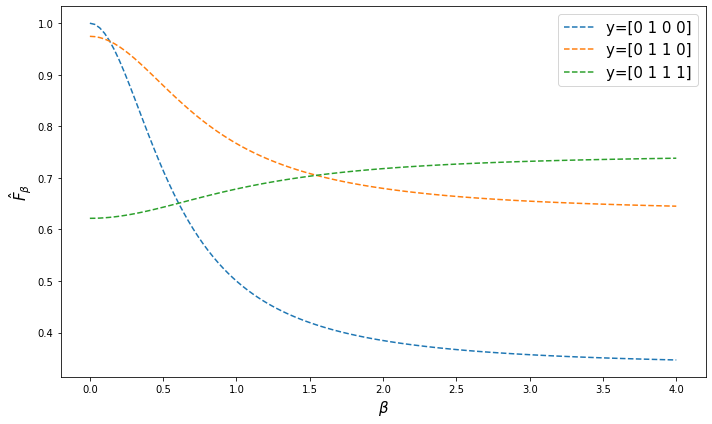}
        \caption{$ \hat{F}_\beta $ for different labeling choices and values of $ \beta $, under the simplified MNIST model. For small $\beta$, the labeling [0,1,0,0] is the best (digit 1 only is anomalous). For intermediate $\beta$, [0,1,1,0] is the best. While for large $\beta$, [0,1,1,1] (all digits are anomalous expect 0) is the best. This simplified result is consistent with our experimental result in~\cref{tab:mnist}.}
        \label{fig:mnist_toy}
    \end{figure}

    We transform each MNIST by randomly cropping it to $25 \times 25$ and rolling between $0$ and $5$ pixels along each dimension independently. We define our models $ A_{\theta_s}, A_{\theta_q} $ as convolutional neural networks (CNNs). As shown in Figure~\ref{fig:mnist_arch}, our networks consist of four 2D convolutions with $ 10 $ out channels, kernel size $ 5 $, padding $ 2 $, and strides $ 1, 1, 2, 2 $ respectively, followed by three FC layers of output size $ 8, 8, 1 $ respectively. All layers but the final layer are followed by a ReLU activation. We train for $3000$ epochs of $1800$ image pairs, with a batch size of $760$, and test by simultaneously feeding images to each network. We use the Adam optimizer~\cite{Paszke19} with a learning rate of $ 10^{-4} $. We set $ \alpha = 0.15 $, $ \lambda_\text{wall} = 1/\alpha $, and $ \lambda_\text{mag} = 0 $. Lastly, we perform early stopping for each run and select the best run from multiple different initial seeds by evaluating our $ \hat{F}_\beta $ metric over the validation set.
    \begin{figure}[htbp]
        \centering
        \scalebox{0.8}{
            \begin{tikzpicture}[node distance=1.1cm]
                \node (data_in) [] {Input};
                \node (conv1) [nn, below of=data_in] {2D Convolution + ReLU};
                \node (conv2) [nn, below of=conv1] {2D Convolution + ReLU};
                \node (conv3) [nn, below of=conv2] {2D Convolution + ReLU};
                \node (conv4) [nn, below of=conv3] {2D Convolution + ReLU};
                \node (fc1) [nn, below of=conv4] {Fully-connected + ReLU};
                \node (fc2) [nn, below of=fc1] {Fully-connected + ReLU};
                \node (fc3) [nn, below of=fc2] {Fully-connected + Sigmoid};
                \node (pred) [below of=fc3] {Anomaly Probability};
        
                \draw [arrow] (data_in) -- (conv1) node [pos=0.5, right] {(1,25,25)};
                \draw [arrow] (conv1) -- (conv2) node [pos=0.5, right] {(10,25,25)};
                \draw [arrow] (conv2) -- (conv3) node [pos=0.5, right] {(10,25,25)};
                \draw [arrow] (conv3) -- (conv4) node [pos=0.5, right] {(10,13,13)};
                \draw [arrow] (conv4) -- (fc1) node [pos=0.5, right] {490};
                \draw [arrow] (fc1) -- (fc2) node [pos=0.5, right] {8};
                \draw [arrow] (fc2) -- (fc3) node [pos=0.5, right] {8};
                \draw [arrow] (fc3) -- (pred) node [pos=0.5, right] {1};
            \end{tikzpicture}
        }
        \caption{\label{fig:mnist_arch} Model architecture for each MNIST DNN, showing the input size to each layer.}
    \end{figure}

    For completeness, \cref{fig:mnist} shows the violin plots corresponding to the results in~\cref{tab:mnist}. We also visualize the latent representations of the digits, generated by the models $ A_{\theta_s} $ and $ A_{\theta_q} $, in~\cref{fig:mnist_latent}.
    \begin{figure}[htbp]
        \centering
        \subfigure{
            \begin{overpic}[width=0.335\linewidth,trim={0 0 0 2.5cm},clip]{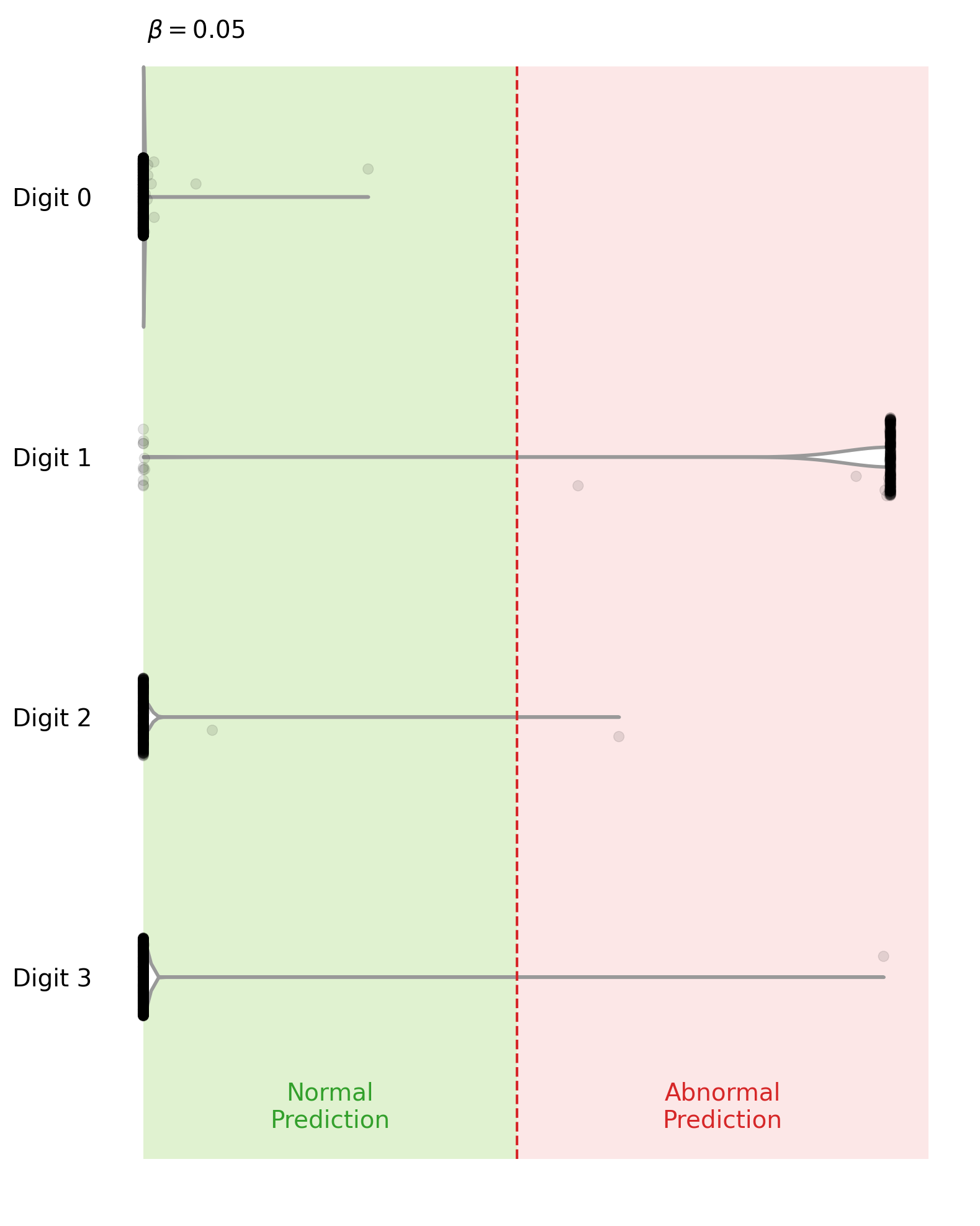}
                \put(54,94){$\beta=0.01$}
            \end{overpic}
        }
        \subfigure{
            \begin{overpic}[width=0.3\linewidth,trim={2cm 0 0 1cm},clip]{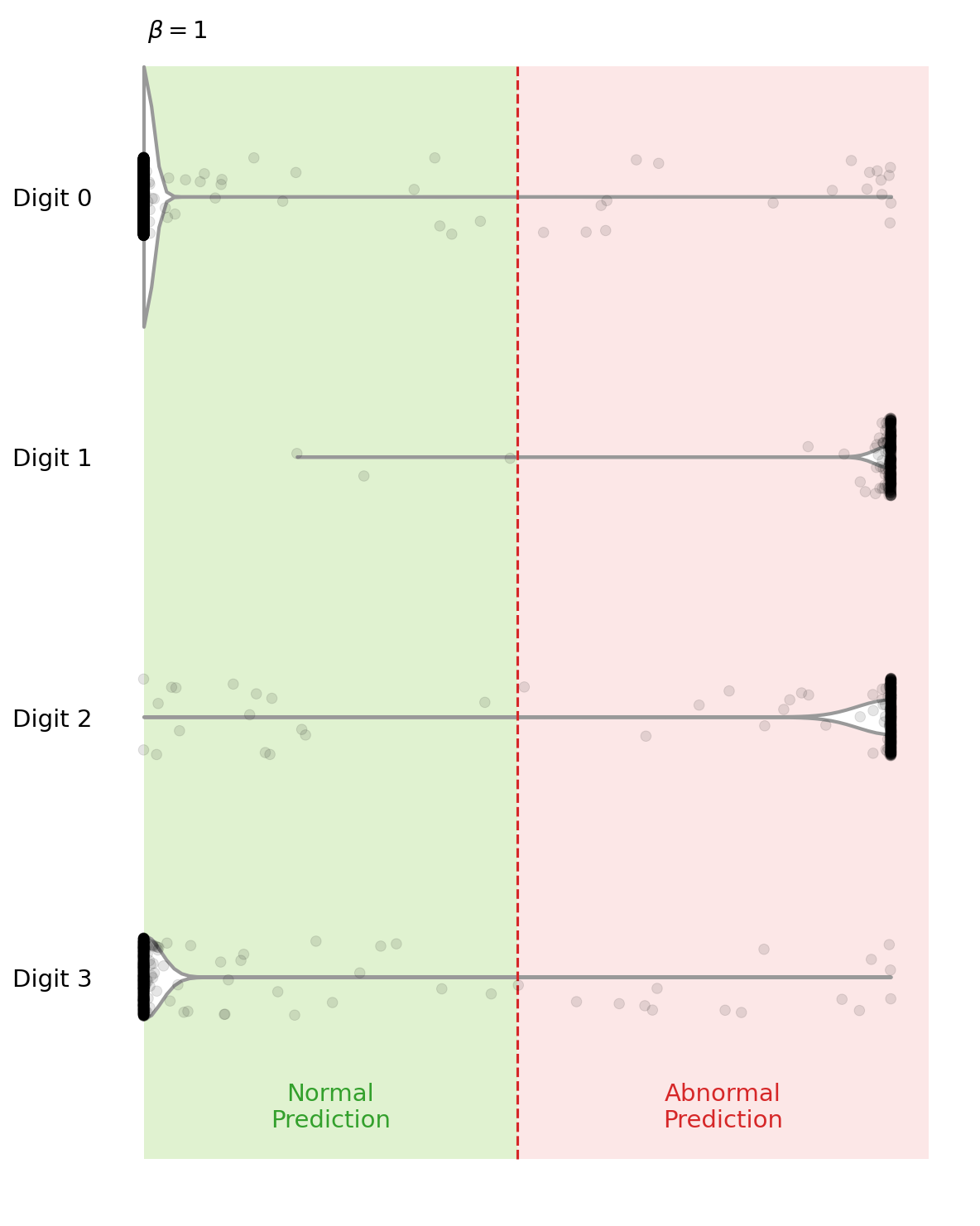}
                \put(52,94){$\beta=1$}
            \end{overpic}
        }
        \subfigure{
            \begin{overpic}[width=0.3\linewidth,trim={2cm 0 0 1cm},clip]{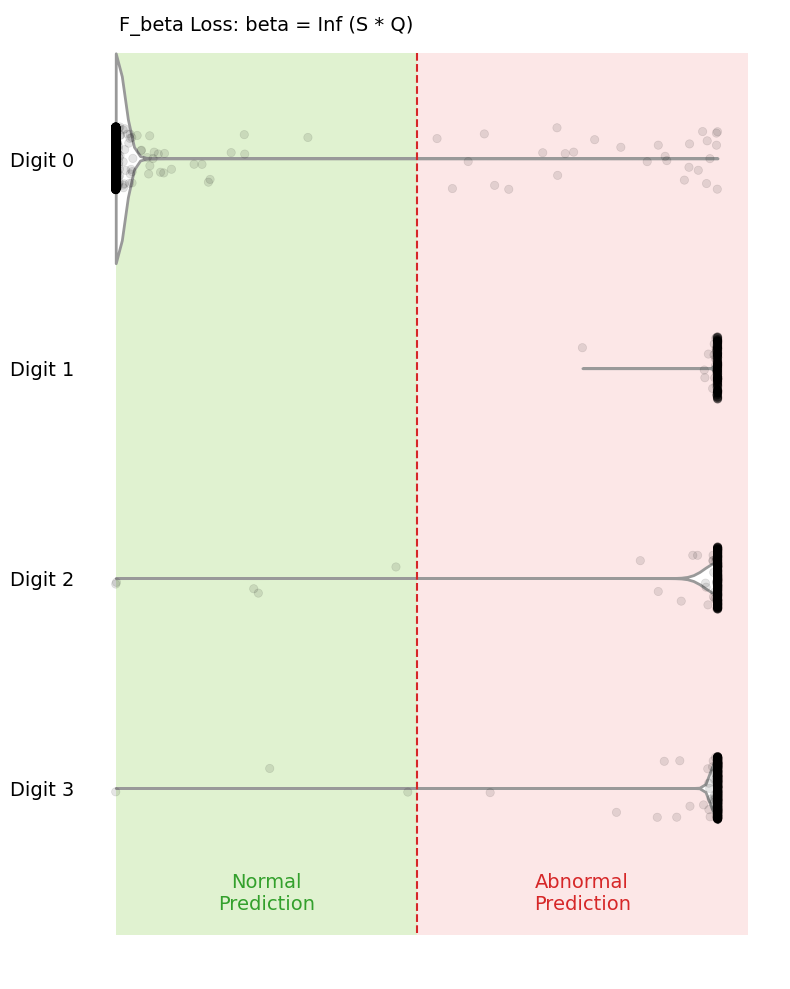}
                \put(50,94){$\beta=\infty$}
            \end{overpic}
        }
        \caption{Violin plot showing predictions from the $\hat{F}_\beta$ models on the MNIST digits. Prediction values are the products of the two network outputs. From left to right, \( \beta = 0.01, 1, \infty \).}
        \label{fig:mnist}
    \end{figure}

    \begin{figure}[htbp]
        \centering
        \subfigure{
            \includegraphics[width=0.3\linewidth,trim={0.2cm 0.2cm 0.2cm 0.88cm},clip]{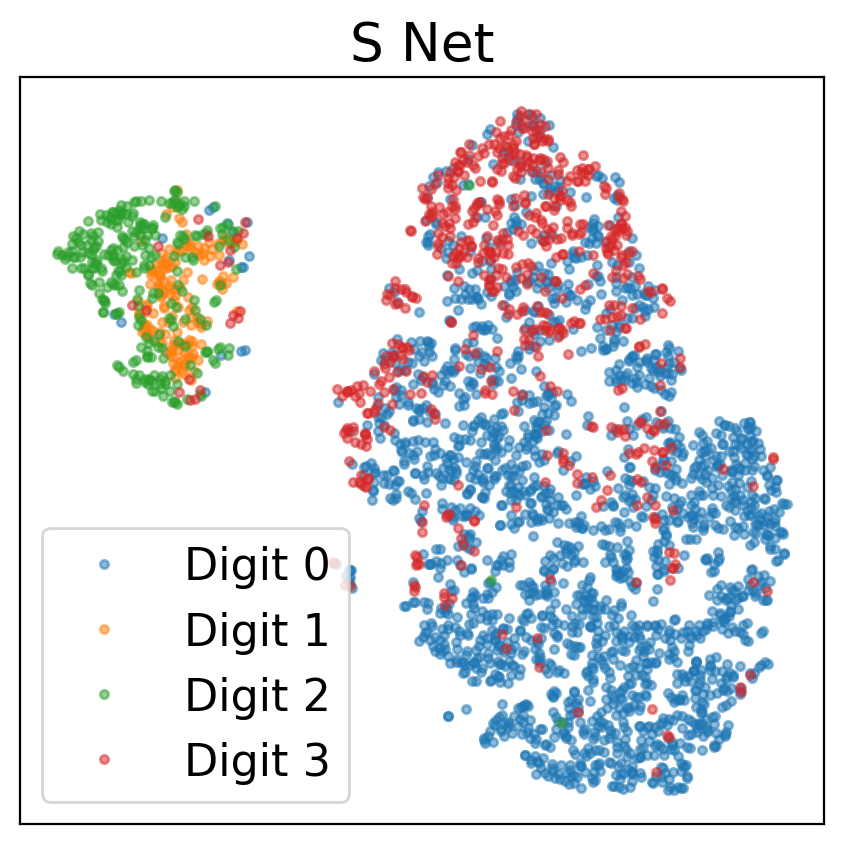}
        }
        \subfigure{
            \includegraphics[width=0.3\linewidth,trim={0.2cm 0.2cm 0.2cm 0.88cm},clip]{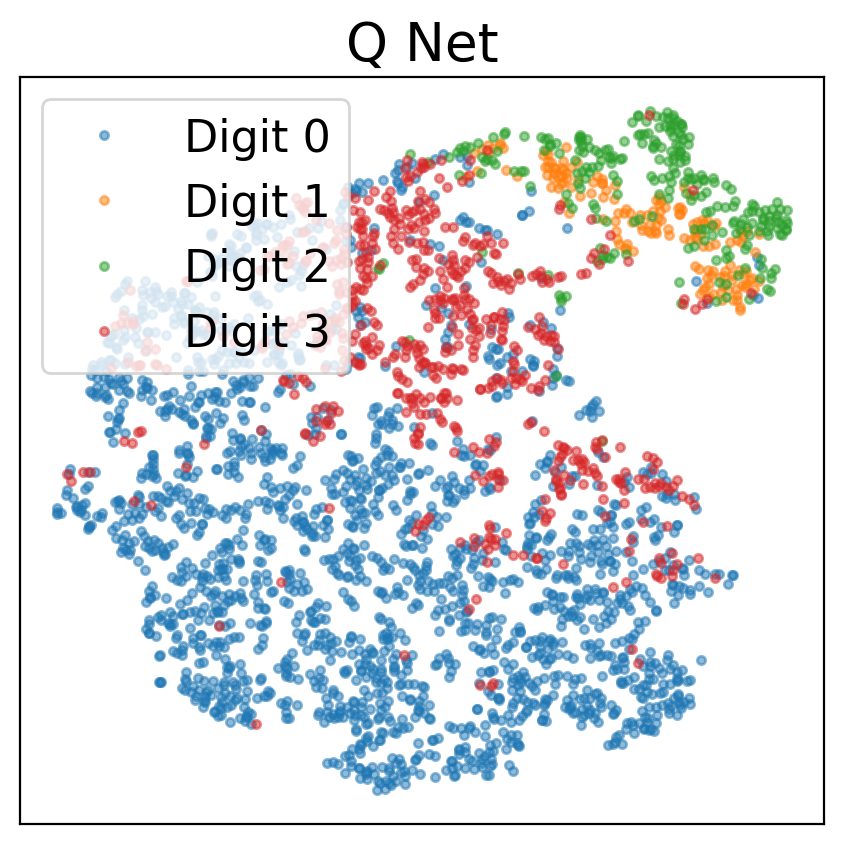}
        }
        \caption{t-SNE~\cite{vandermaaten08} visualization of the latent representations of the MNIST digits. We show the representations of $ A_{\theta_s} $ (left) and $ A_{\theta_q}$ (right) for $ \beta = 1 $.}
        \label{fig:mnist_latent}
    \end{figure}
\end{toappendix}

\subsection{Metal milling dataset}\label{sec:milling}
We now assess our method on a real-world dataset. The University of California, Berkeley Milling dataset~\cite{agogino07g} is an open dataset of acoustic, vibration, and current measurements from a set of metal milling cuts. A recent paper~\cite{hahn21m} provides a detailed description of the task, analysis code, and results from a variational autoencoder (VAE). The dataset consists of 167 different milling cuts, corresponding to a total of approximately 100 minutes of milling. There are six total diagnostics: acoustics and vibrations from the spindle, acoustics and vibration from the table, and AC and DC current. In addition, the degree of flank wear on the milling tool is measured after a selection of the cuts. We follow the task as described in~\cite{hahn21m}, breaking the data into 0.25 second chunks, and try to predict whether the milling performance in each chunk is “healthy” or “degraded/failed,” with the label determined by the degree of flank wear.

\begin{figure}[htbp]
    \centering
    \begin{minipage}{0.66\textwidth}
    \subfigure{
        \begin{overpic}[width=0.51\linewidth,trim={0.2cm 0.8cm 0.8cm 0.8cm},clip]{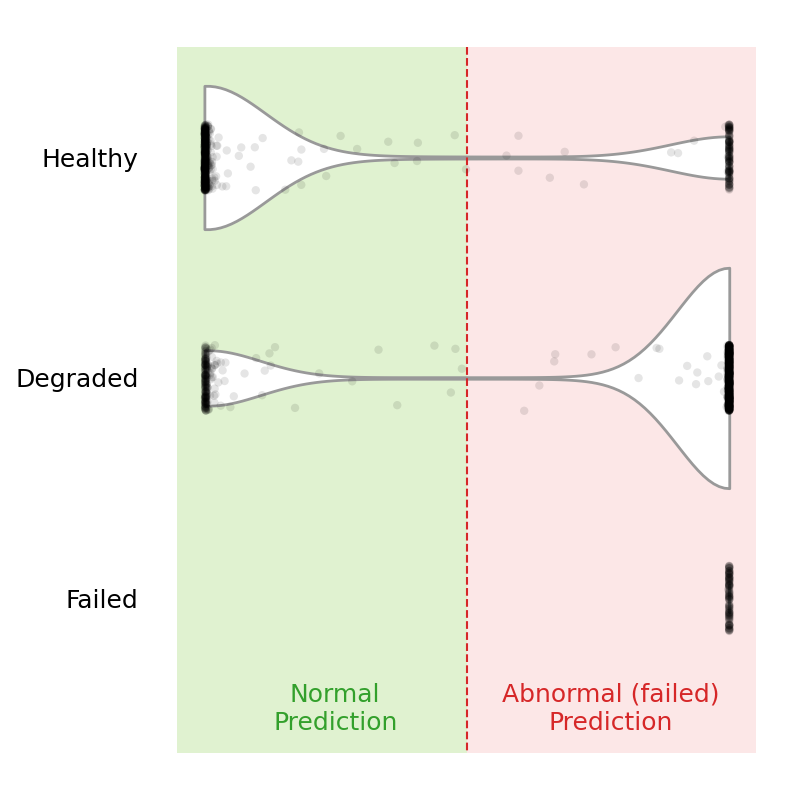}
            \put(76,88){CoAD}
        \end{overpic}
    }
    \subfigure{
        \begin{overpic}[width=0.41\linewidth,trim={3cm 0 0 0},clip]{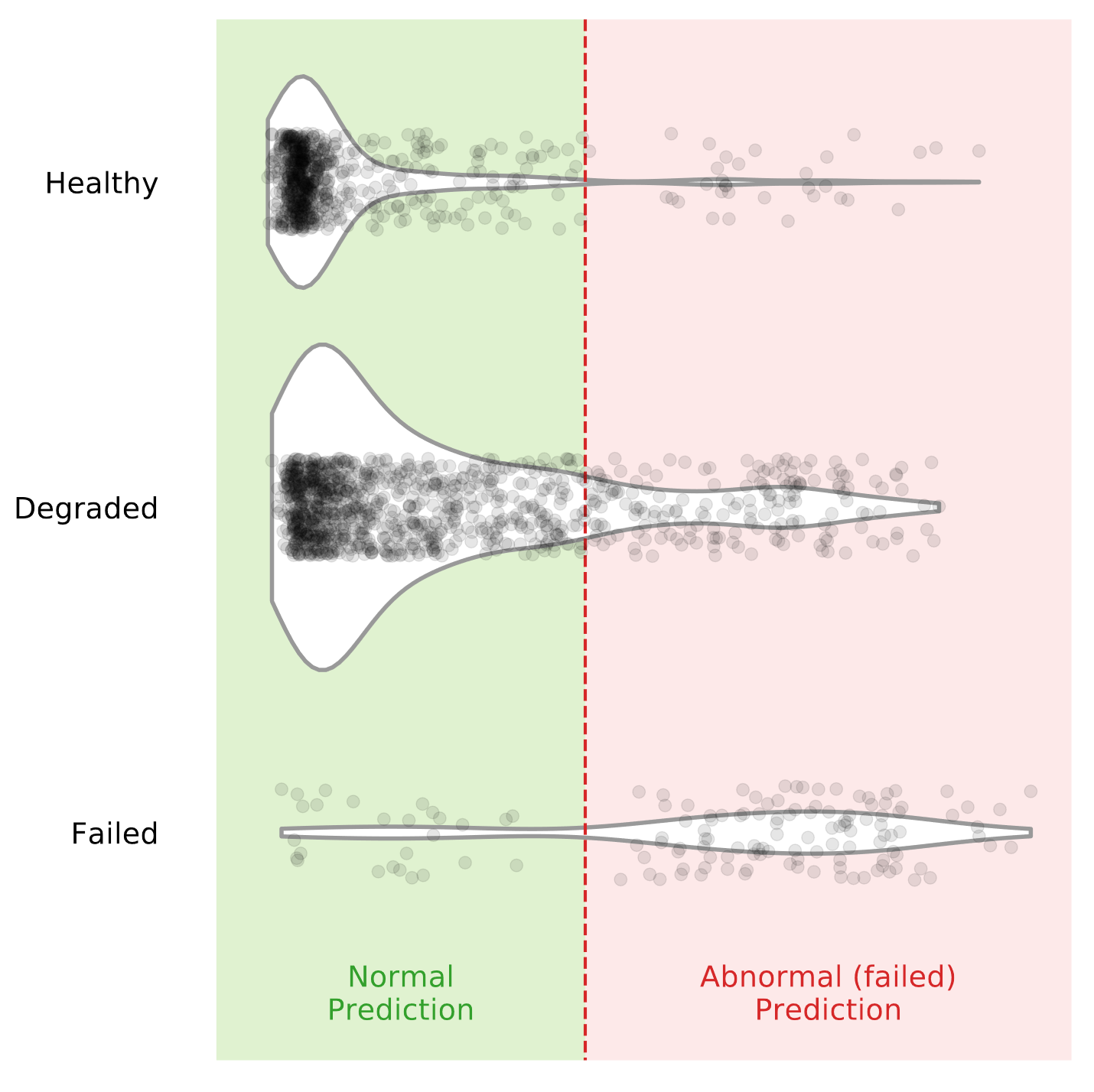}
            \put(64,90){VAE}
        \end{overpic}
    }
    \end{minipage}\hspace{1em}
    \begin{minipage}{0.3\textwidth}
    \caption{Milling data set results. (Left) Violin plot showing predictions from the $\hat{F}_{10}$ model; the prediction values are the products of the two network outputs. (Right) Violin plot showing predictions from the VAE in~\cite{hahn21m}. }
    \label{fig:milling_pred}
    \end{minipage}
\end{figure}

To apply coincident learning to the milling task, we divide the diagnostics into two sets: acoustics and vibration measurements (four data inputs) and AC/DC current measurements (two data inputs). (Note that in the milling dataset there is not a default separation into “subsystem” and “quality” measurements.) Also, in part because the “degraded/failed” examples are the majority case, our algorithm learns to identify the most “healthy” examples as a distinct class.
\cref{fig:milling_pred} shows the predictions from the networks trained under coincident learning (at \( \beta = 10 \)) as compared to the predictions from the VAE of~\cite{hahn21m}. Our model has perfect predictions on failed examples while identifying more “healthy” and “degraded” examples as anomalous compared to the VAE. We emphasize that in contrast to the VAE, the coincident model does not require training only on “healthy” data. Lastly, as shown in~\cref{tab:milling-benchmark}, CoAD achieves a better $F_1$ score (where we consider “degraded/failed” as abnormal) than the VAE and several other deep anomaly detection methods.

\begin{table}[htbp]
    \caption{Comparison of $F_1$ scores on metal milling dataset. VAE trained on only “healthy” data.}
    \label{tab:milling-benchmark}
    \centering
    \begin{tabular}{cccc}
    \toprule
    VAE~\cite{hahn21m} & DGHL~\cite{challu2022deep} & OmniAnomaly~\cite{su2019robust} & CoAD \\ \midrule
    0.79 & 0.80 & 0.80 & \textbf{0.86} \\ \bottomrule
    \end{tabular}
\end{table}

\begin{figure}[htbp]
    \centering
    \begin{minipage}{0.36\linewidth}
    \includegraphics[width=\linewidth]{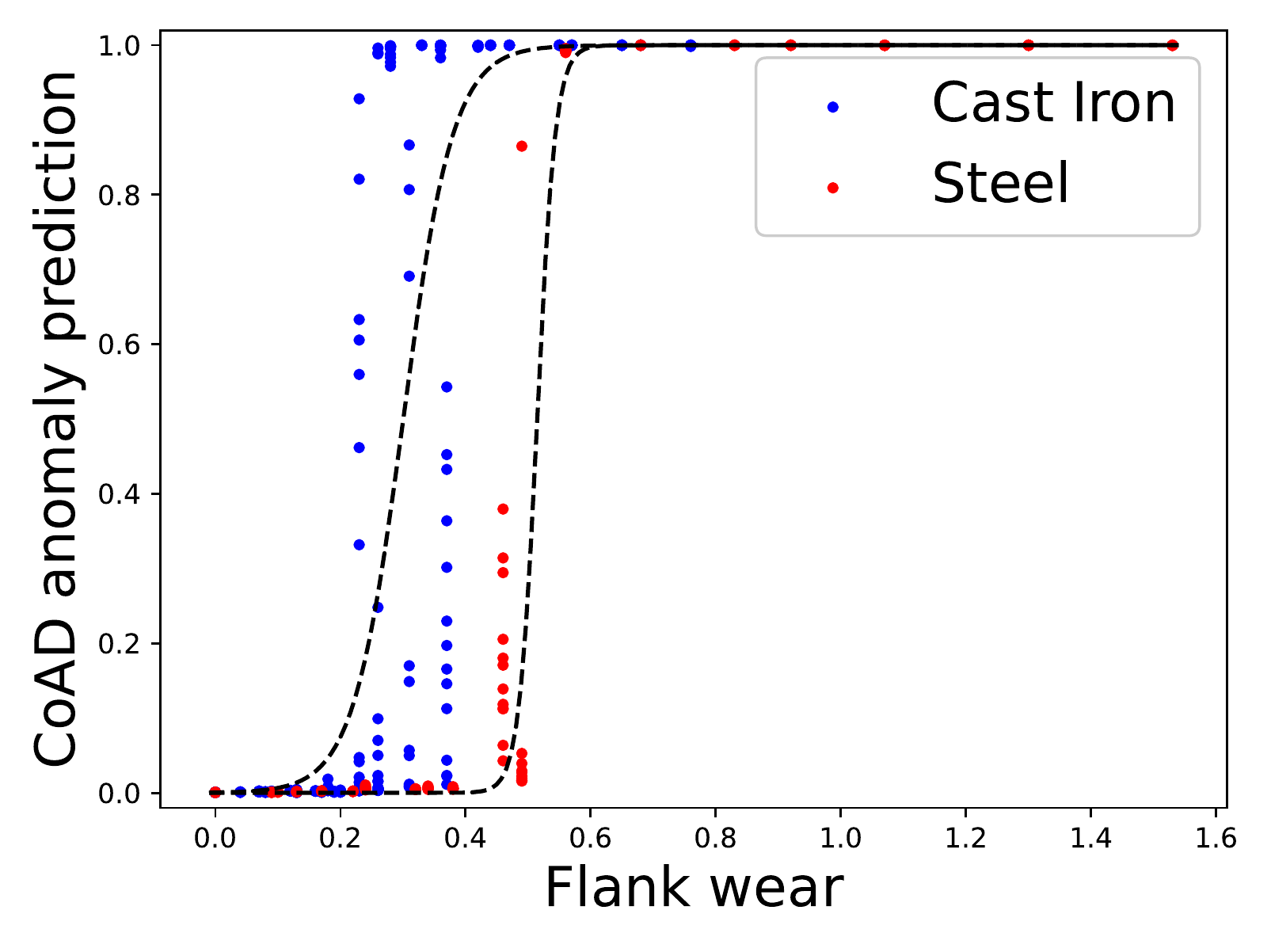}
    \end{minipage}\hspace{1em}
    \begin{minipage}{0.54\linewidth} 
    \caption{CoAD anomaly prediction versus flank wear for two different milling configurations: cast iron (left) and steel (right) at $0.75$mm cut depth and $0.25$mm/rev cut speed. The dashed lines are sigmoid fits for illustration.}
    \label{fig:milling_flankwear_slim}
    \end{minipage}
\end{figure}

The labels defined in~\cite{hahn21m} are quite simplistic, using only the degree of flank wear and ignoring the other milling parameters (metal type, cut speed, and cut depth). While the flank wear is indicative of milling performance, the amount of wear that will lead to anomalous milling may differ across these different milling settings. \cref{fig:milling_flankwear_slim} shows the model's anomaly confidence versus the flank wear for two different milling configurations (the other six are shown in the Appendix). There is only a weak correlation between our predictions and flank wear in \textit{aggregate} across all eight milling settings but a very strong correlation for each \textit{individual} configuration. Labels based on flank wear alone would therefore be incorrect. 
We emphasize that the models never see the flank wear measurements or milling configurations, and are trained simultaneously on all milling configurations.

\begin{toappendix}
    \subsection{Milling dataset}

    We define our models $ A_{\theta_s}, A_{\theta_q} $ as CNNs. As shown in Figure~\ref{fig:milling_arch}, our networks consist of two 1D convolutions with $ 10 $ out channels, kernel size $ 5 $, padding $ 2 $, and stride $ 1 $, followed by three FC layers of output size $ 8, 8, 1 $ respectively. All layers but the final layer are followed by a ReLU activation. The $ s $ net has $ 2 $ input channels; the $ q $ net has $ 4 $ input channels. We train for $150$ epochs using a batch size of $1024$ and the train-test splits of~\cite{hahn21m}. We use the Adam optimizer~\cite{Paszke19} with a learning rate of $ 3 \times 10^{-3} $. We set $ \alpha = 0.366 $, according to the flank wear labels. We set $ \lambda_\text{wall} = 1 $ and $ \lambda_\text{mag} = 2e-4 $.
    \begin{figure}[htbp]
        \centering
        \scalebox{0.8}{
            \begin{tikzpicture}[node distance=1.1cm]
                \node (data_in) [] {Input};
                \node (conv1) [nn, below of=data_in] {1D Convolution + ReLU};
                \node (conv2) [nn, below of=conv1] {1D Convolution + ReLU};
                \node (fc1) [nn, below of=conv2] {Fully-connected + ReLU};
                \node (fc2) [nn, below of=fc1] {Fully-connected + ReLU};
                \node (fc3) [nn, below of=fc2] {Fully-connected + Sigmoid};
                \node (pred) [below of=fc3] {Anomaly Probability};
        
                \draw [arrow] (data_in) -- (conv1) node [pos=0.5, right] {(2 or 4,64)};
                \draw [arrow] (conv1) -- (conv2) node [pos=0.5, right] {(10,64)};
                \draw [arrow] (conv2) -- (fc1) node [pos=0.5, right] {640};
                \draw [arrow] (fc1) -- (fc2) node [pos=0.5, right] {8};
                \draw [arrow] (fc2) -- (fc3) node [pos=0.5, right] {8};
                \draw [arrow] (fc3) -- (pred) node [pos=0.5, right] {1};
            \end{tikzpicture}
        }
        \caption{\label{fig:milling_arch} Model architecture for the milling DNNs, showing the input size to each layer.}
    \end{figure}

    For the comparison to DGHL~\cite{challu2022deep} and OmniAnomaly~\cite{su2019robust}, we leverage their public repositories: \href{https://github.com/cchallu/dghl}{DGHL} and \href{https://github.com/NetManAIOps/OmniAnomaly}{OmniAnomaly}. For DGHL, we used the default training parameters and architecture, except for setting the feature size to $6$, learning rate to $10^{-4}$, and iteration steps to $10^4$. For OmniAnomaly, we used the default training parameters and architecture, except for setting the window length to $64$. Due to an issue with the public repository, we ran on a reduce random subset of the training data to fit within GPU memory and avoid OOM errors.

    As discussed in~\cref{sec:milling}, the labels defined in~\cite{hahn21m} are quite simplistic, since they ignore the milling parameters: metal type (iron or steel), cut speed (0.25mm/rev or 0.5mm/rev), and cut depth (0.75mm or 1.5mm). \cref{fig:milling_flankwear} shows the model's anomaly confidence versus the flank wear for the 8 different milling configurations (using $ \beta = 10 $). We see there is, at best, a weak correlation between our predictions and flank wear in \textit{aggregate} but a very strong correlation for each \textit{individual} configuration. In fact, our anomaly confidence is physically interpretable: as the feed rate and depth increase (i.e., more material volume cut per revolution), the degree of flank wear corresponding to anomalous milling decreases.

    
    \begin{figure}[htbp]
        \centering
        \subfigure{
            \includegraphics[width=0.4\linewidth]{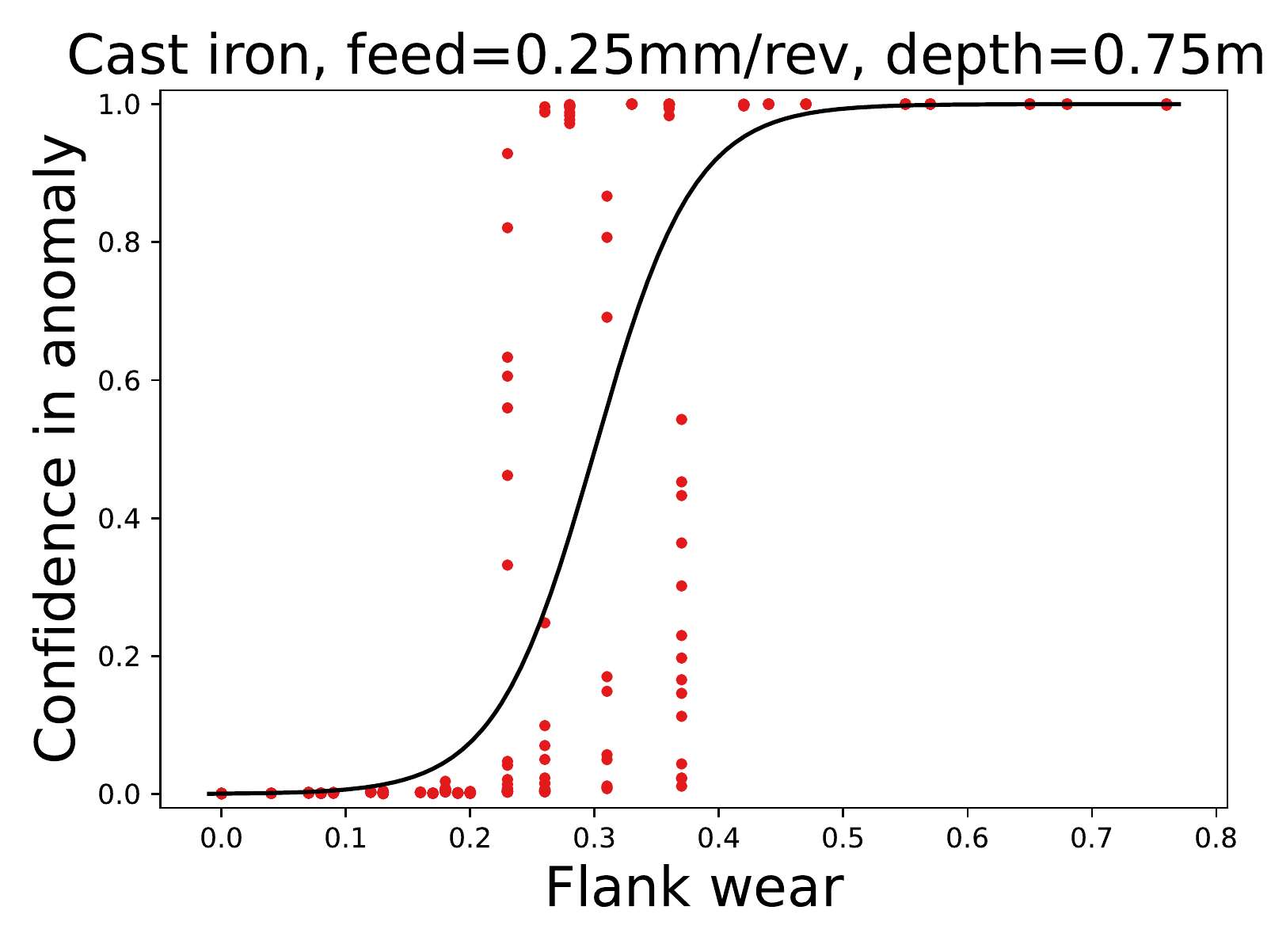}
        }
        \subfigure{
            \includegraphics[width=0.4\linewidth]{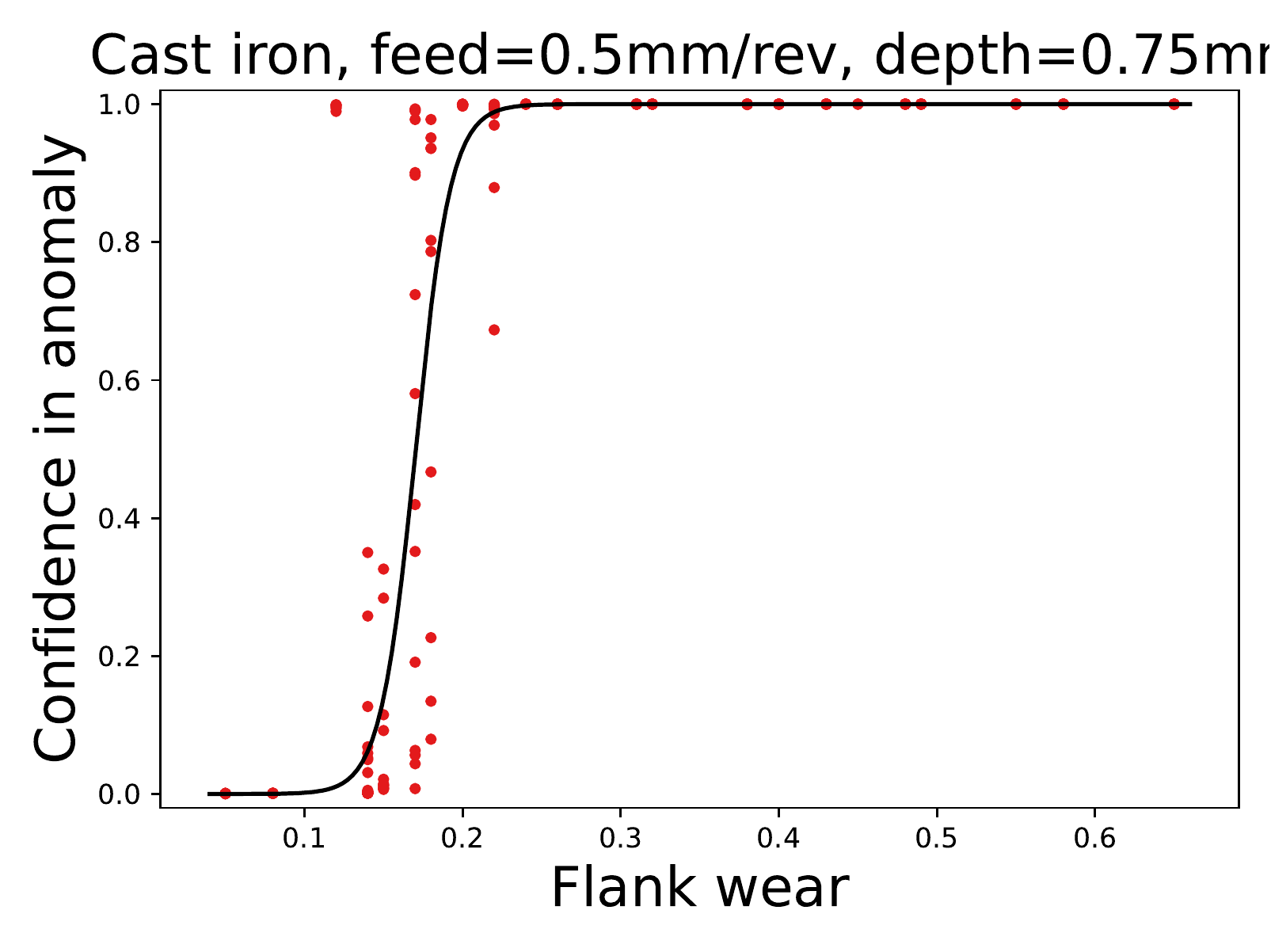}
        }
        \subfigure{
            \includegraphics[width=0.4\linewidth]{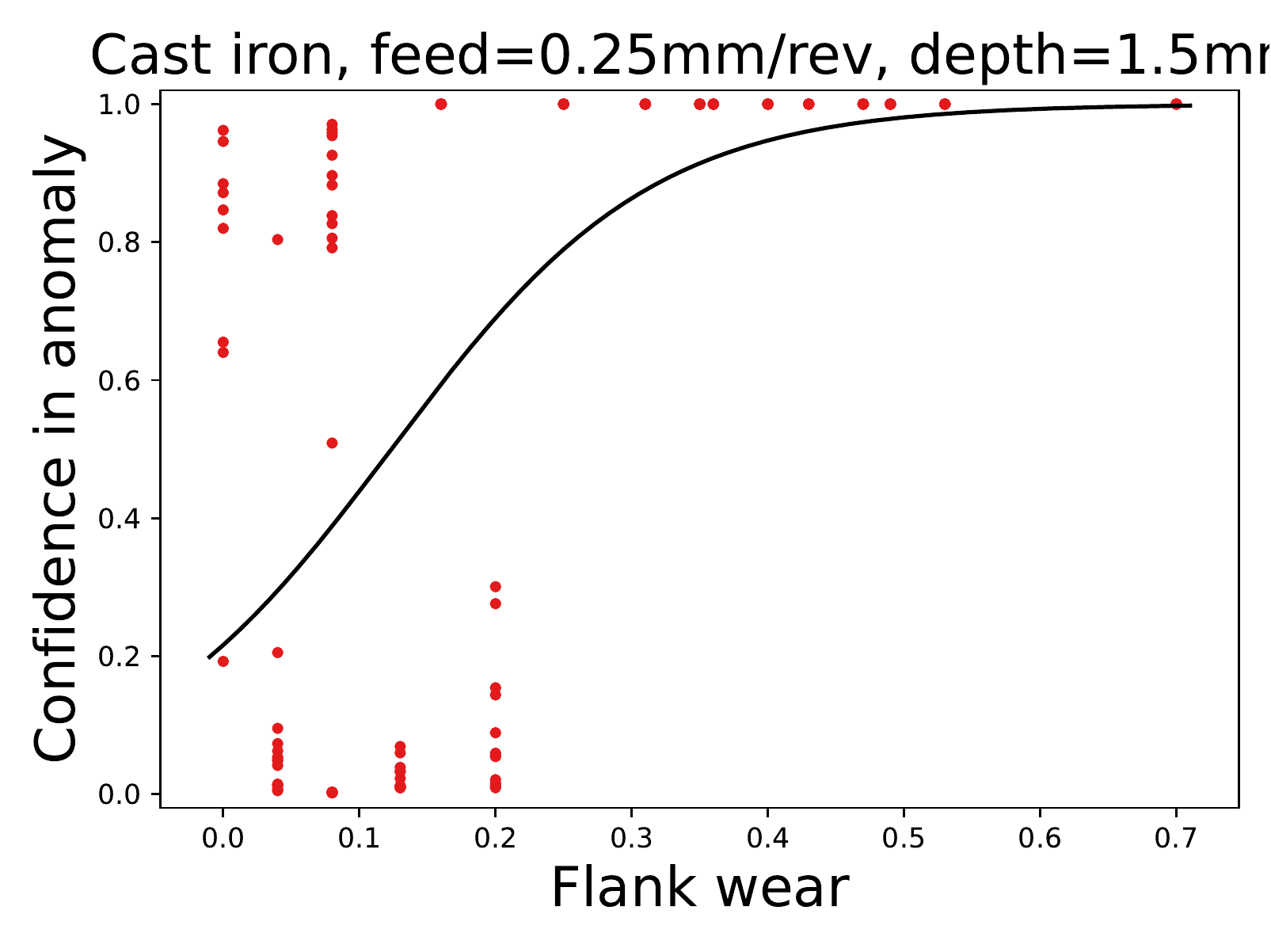}
        }
        \subfigure{
            \includegraphics[width=0.4\linewidth]{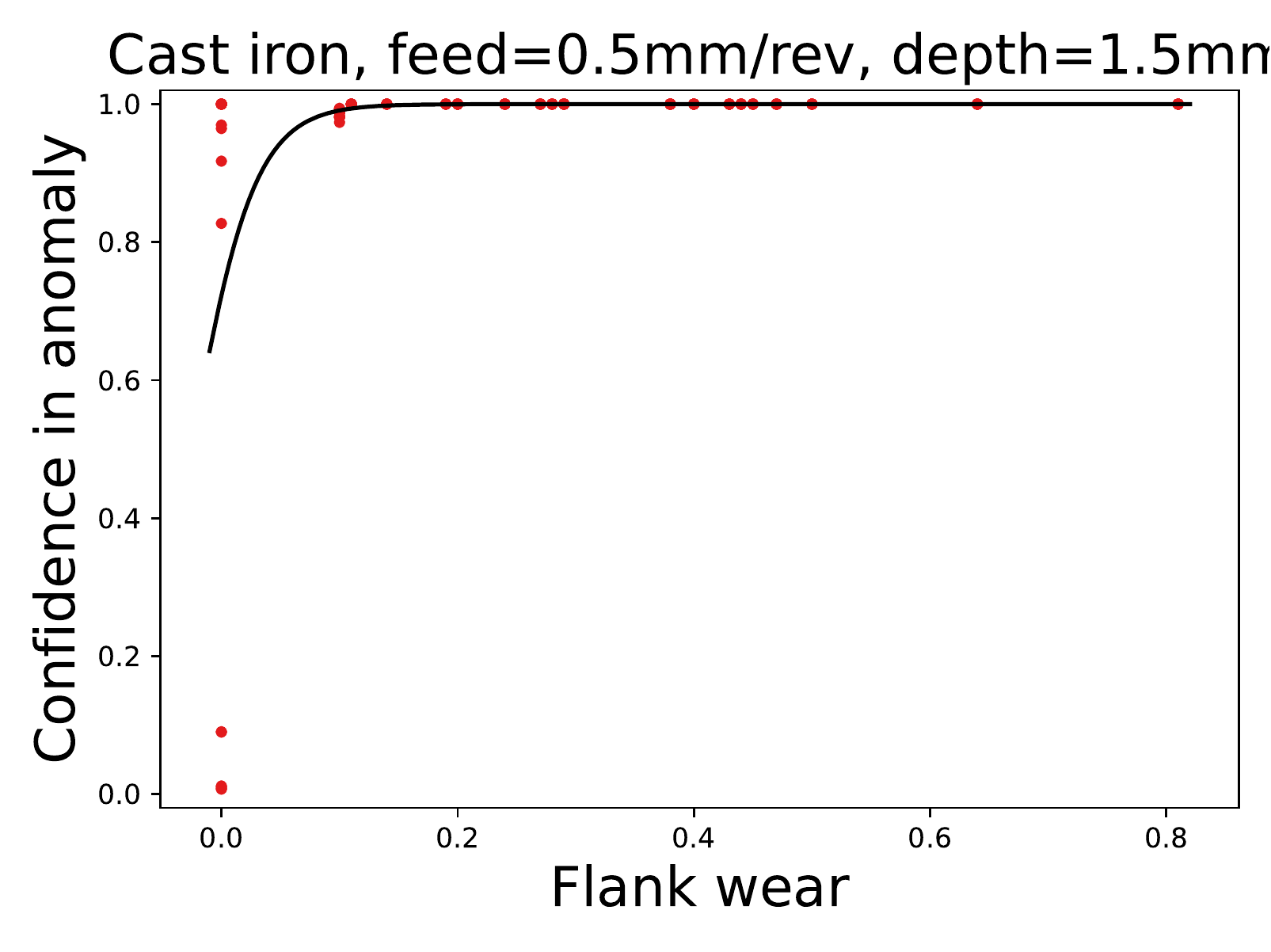}
        }
        \subfigure{
            \includegraphics[width=0.4\linewidth]{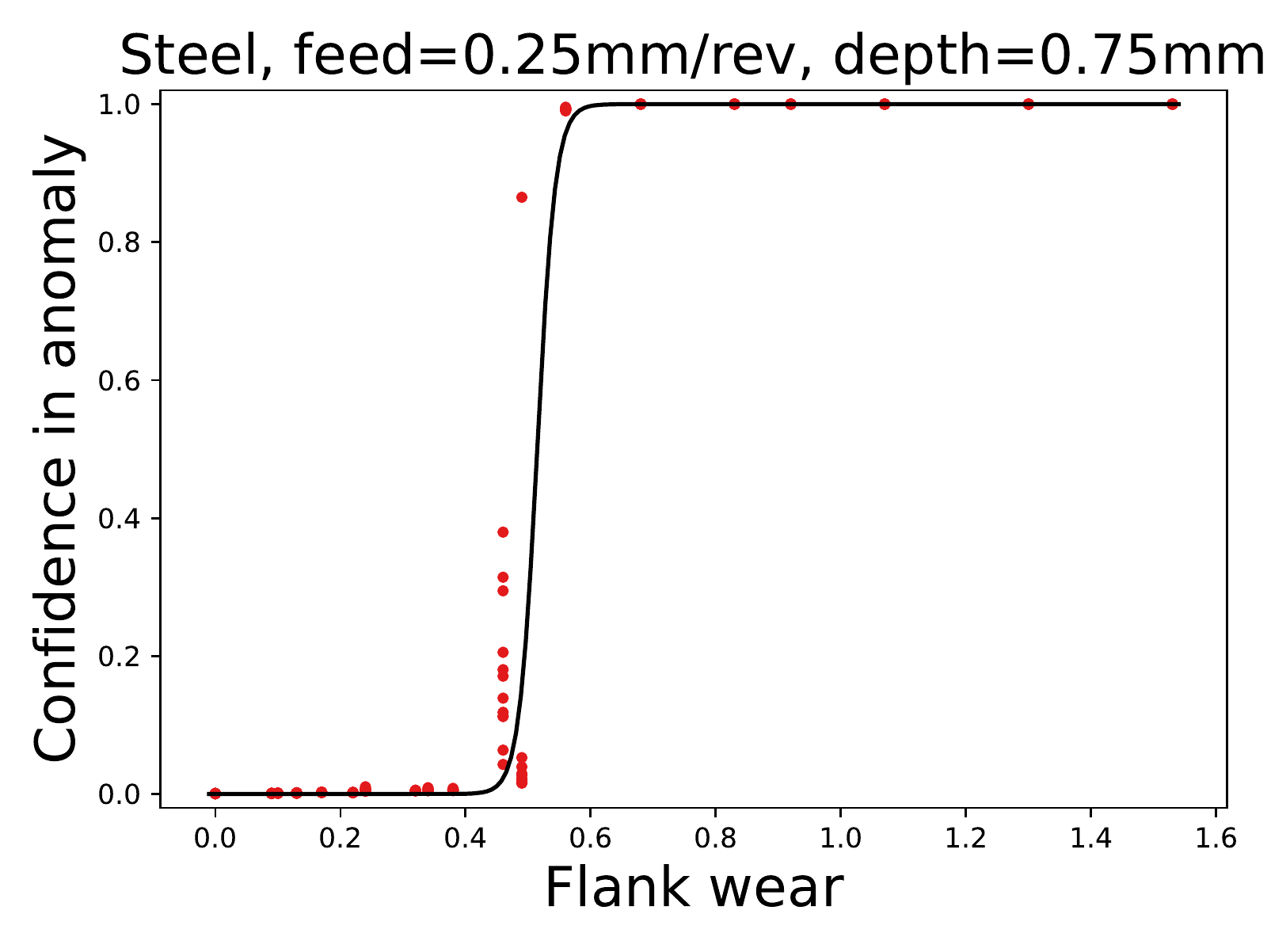}
        }
        \subfigure{
            \includegraphics[width=0.4\linewidth]{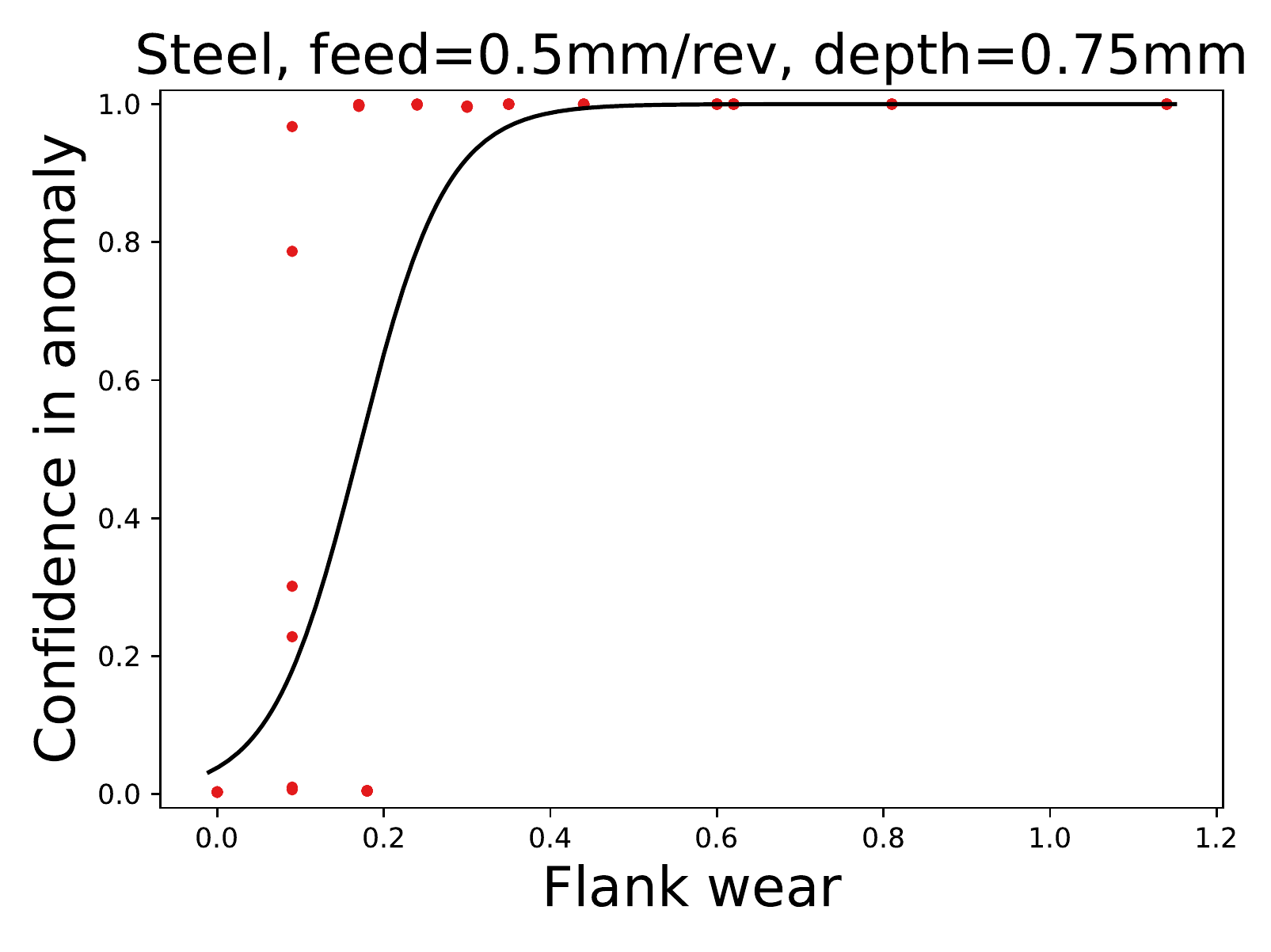}
        }
        \subfigure{
            \includegraphics[width=0.4\linewidth]{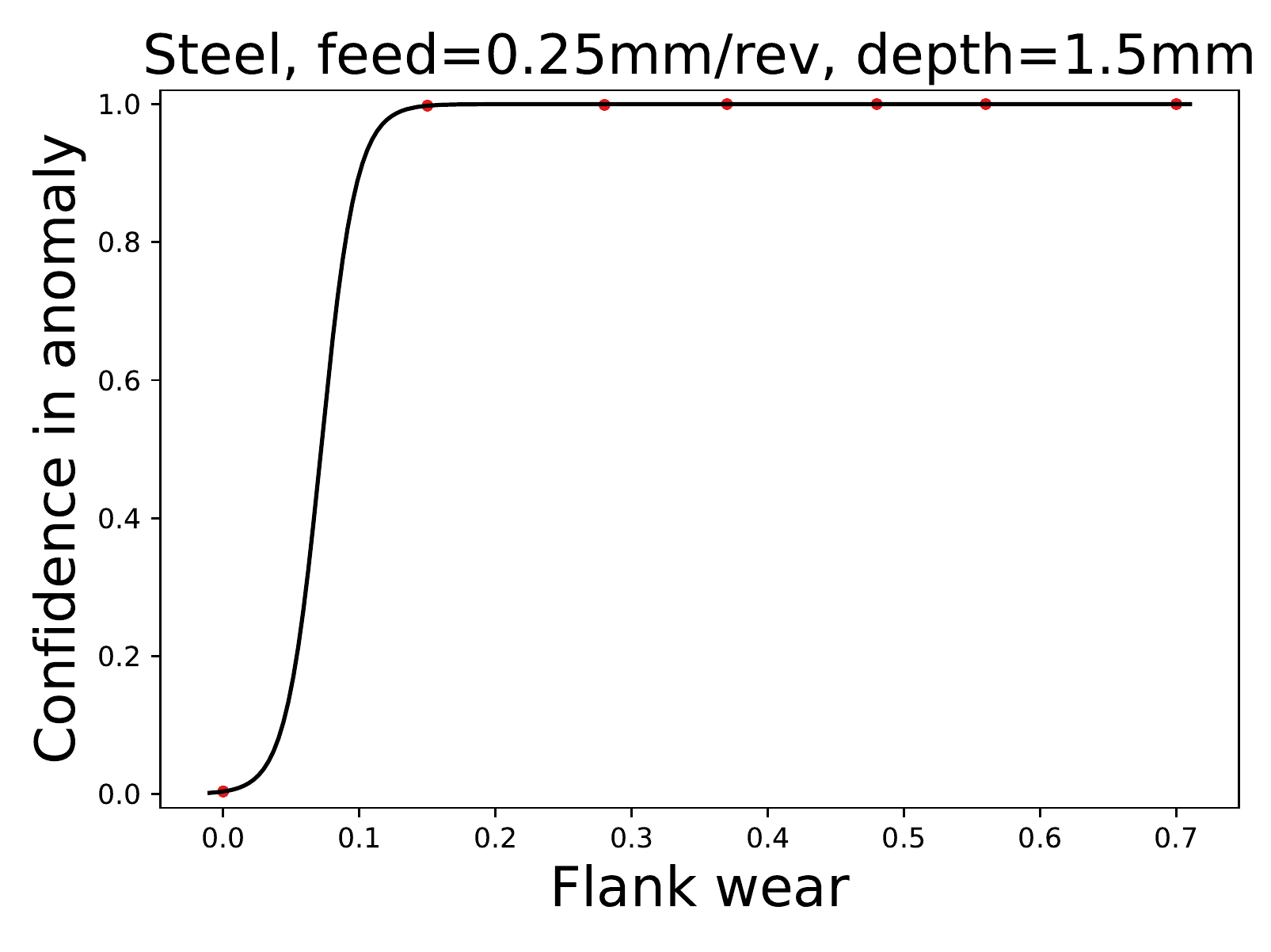}
        }
        \subfigure{
            \includegraphics[width=0.4\linewidth]{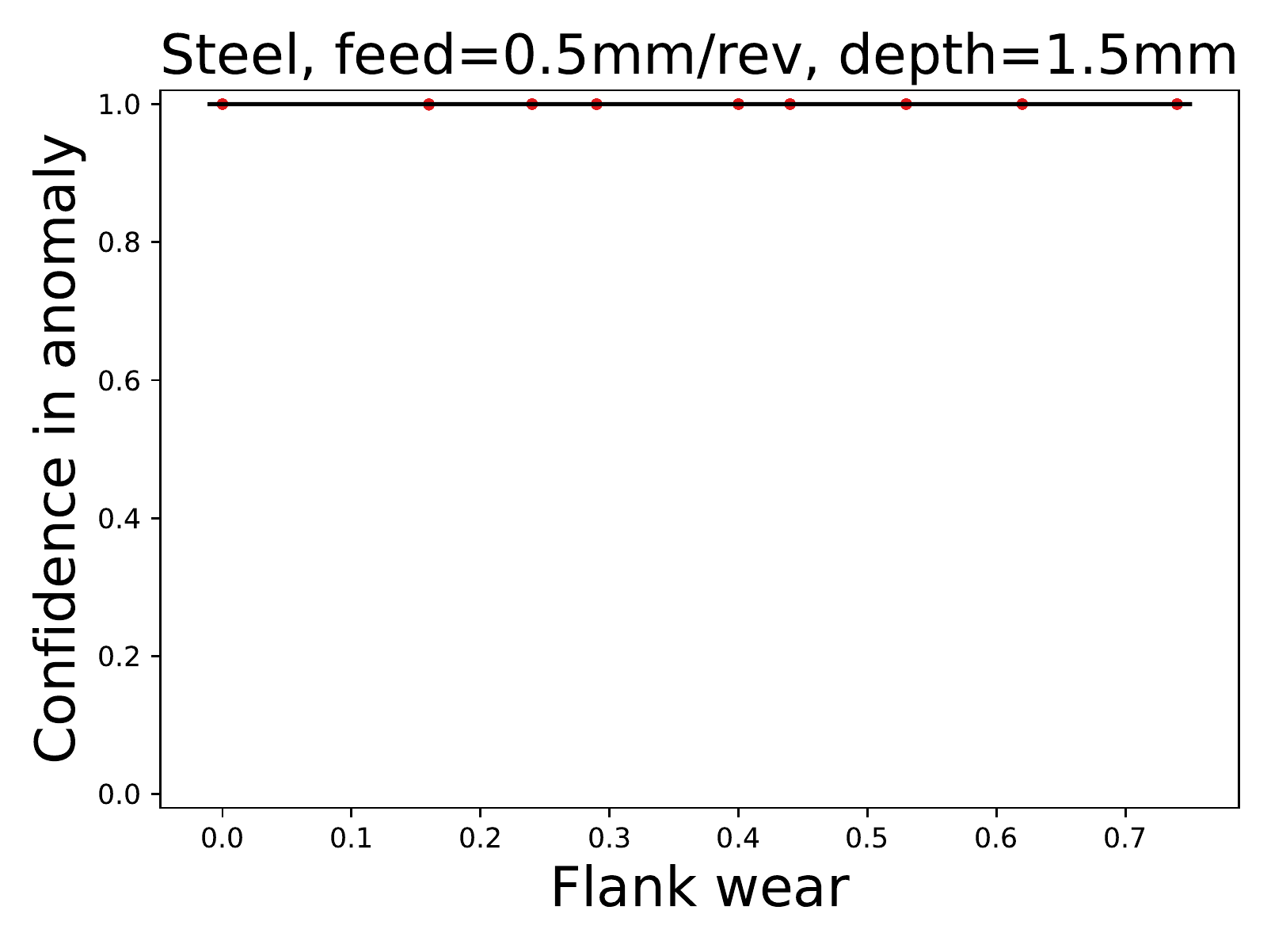}
        }
        \caption{Anomaly probability versus flank wear for different milling configurations. Black line is a sigmoid fit.}
        \label{fig:milling_flankwear}
    \end{figure}
\end{toappendix}

\subsection{Particle accelerator dataset}\label{sec:particle_accel}
We now return to our original motivating problem of identifying the source of RF station faults. We utilize the dataset assembled and described in~\cite{Humble2022}. The subsystem ($s$) data input consists of time-series data for a single RF station, with one data point approximately every 5 seconds. We use a sensitive trigger to actively select time windows with the possibility of an event to reduce data requirements; any relative change of 0.5\% will trigger a window to be acquired. The quality ($q$) data input consists of beam-position monitor (BPM) data from seven different BPM diagnostics in dispersive (i.e., energy-sensitive) regions of the accelerator. Each BPM input consists of beam positions in the dispersive direction, recorded synchronously at 120 Hz. Time windows are selected to cover the 8 seconds prior to the end of each (asynchronous) RF station event.

\begin{figure}[htbp]
    \centering
    \subfigure[Normal examples]{
        \includegraphics[width=.47\linewidth,trim={0 0.1cm 0 0.9cm},clip,grid]{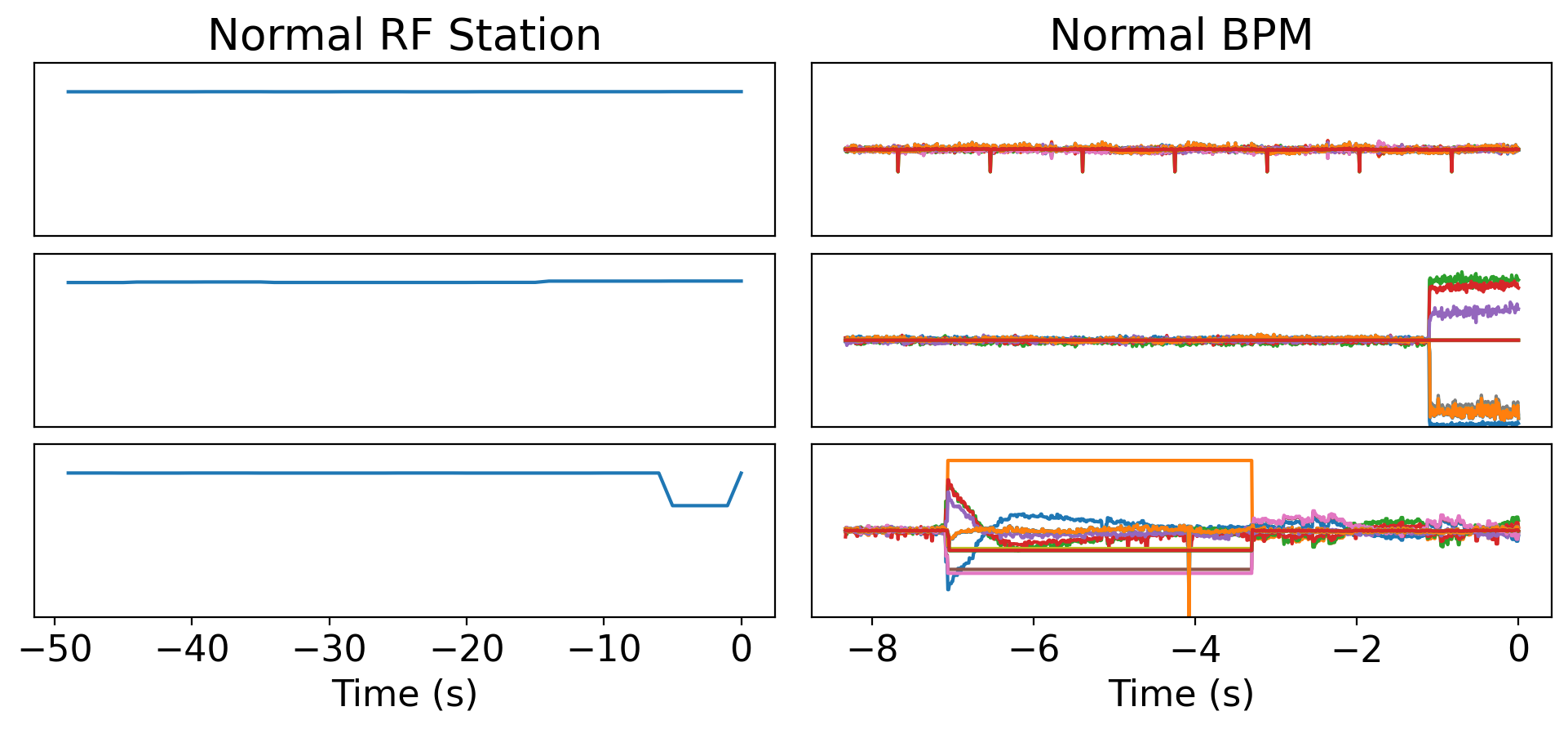}
    }\quad
    \subfigure[Anomalous examples]{
        \includegraphics[width=.47\linewidth,trim={0 0.1cm 0 0.9cm},clip]{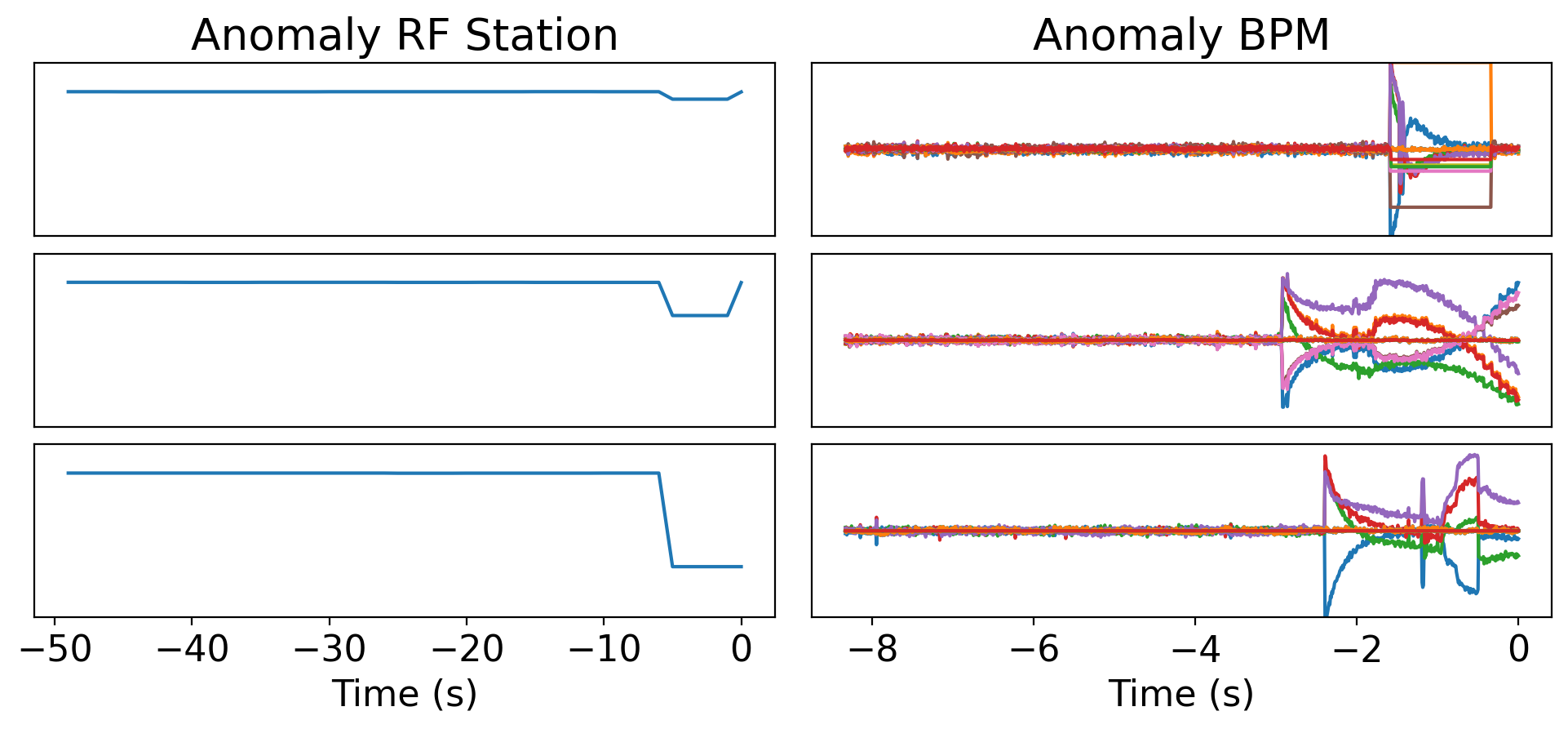}
    }
    \caption{Experimental RF station data examples. Each example shows both the RF station (left) and electron BPM (right) measurement data.}
    \label{fig:rfstation-examples}
\end{figure}

\cref{fig:rfstation-examples} shows examples identified as normal and anomalous. Many of the anomalous cases are apparent by eye. For the normal cases, we specifically selected non-trivial examples that a non-expert might identify as abnormal, but which upon close examination does not correspond to an anomaly in the selected RF station data. For example, the final normal case even contains an abnormal energy deviation, but the algorithm correctly determines that the BPM abnormality is too early compared to the RF station anomaly (as described in~\cite{Humble2022}). With significant effort and manual inspection, it is possible to expertly design anomaly detection models for each data input~\cite{Humble2022}. However, CoAD only has a small set of hyperparameters and avoids the need for a hand-designed anomaly detection algorithm. For accelerators, our approach can scale to cover thousands of potential anomaly sources.

Lastly, we compare CoAD against the prior work in~\cite{Humble2022} and several deep anomaly detection methods, as shown in~\cref{tab:rf-benchmark}. In order to benchmark against methods not designed for multiple inputs, we attempted two strategies: \textit{Score\&Stack} (train and score on each input independently and use the Pareto frontier of all threshold pairs) and \textit{Stack\&Score} (train and score on the stacked input). For these methods, we run both strategies and show the score of whichever approach performed better. CoAD achieves near supervised-level performance (within the expected error rate of the hand labels) and outperforms the other unsupervised methods. We also highlight that prior works require labels to select effective thresholds; amongst these methods, only CoAD (as shown in~\cref{sec:thresh_results}) can select good thresholds without labels.

\begin{table}[htbp]
    \caption{Comparison of $F_1$ scores on RF station anomaly dataset. The expert-designed (Expert) method and supervised classifier (SNN) results are colored in gray as they require supervision.}
    \label{tab:rf-benchmark}
    \centering
    \begin{tabular}{cccccc}
    \toprule
    \textcolor{gray}{Expert}~\cite{Humble2022} & \textcolor{gray}{SNN}~\cite{Humble2022} & OCSVM~\cite{Scholkopf01} & DGHL~\cite{challu2022deep} & OmniAnomaly~\cite{su2019robust} & CoAD \\ \midrule
    \textcolor{gray}{0.90} & \textcolor{gray}{0.88} & 0.66 & 0.64 & 0.80 & \textbf{0.86} \\ \bottomrule
    \end{tabular}
\end{table}

\begin{toappendix}
    \subsection{Particle accelerator RF stations}
    We use the dataset constructed in~\cite{Humble2022} with additional transformations: (i) normalize the data, (ii) add $ 0.1 \times \mathcal{N}\left(0,1\right) $ noise to each input, (iii) roll the BPM data between $0$ and $10$ points randomly, and (iv) select only the last $ 50 $ points in the RF data and the last $ 1000$ in the BPM data. We define our models $ A_{\theta_s}, A_{\theta_q} $ as CNNs. As shown in~\cref{fig:rf_arch_s}, our $ s $ network consists of two 1D convolutions with $ 5 $ out channels, kernel size $ 10 $, padding $ 0 $, and stride $ 1 $, followed by three FC layers of output size $ 6, 6, 1 $ respectively. As shown in~\cref{fig:rf_arch_q}, our $ q $ network is very similar, using $15$ out channels, a kernel size $ 20 $, and a stride of $ 5 $ instead. All layers but the final layer are followed by a ReLU activation. We train for $2000$ epochs using a batch size of $400$ and a train-test split of $85\%-15\%$. We use the Adam optimizer~\cite{Paszke19} with a learning rate of $ 10^{-4} $. We set $ \alpha = 0.2 $, which is the approximate anomaly rate based on the labels. We set $ \lambda_\text{wall} = 1/\alpha $ and $ \lambda_\text{mag} = 0 $. Lastly, we perform early stopping using our $ \hat{F}_\beta $ metric on the validation set.

    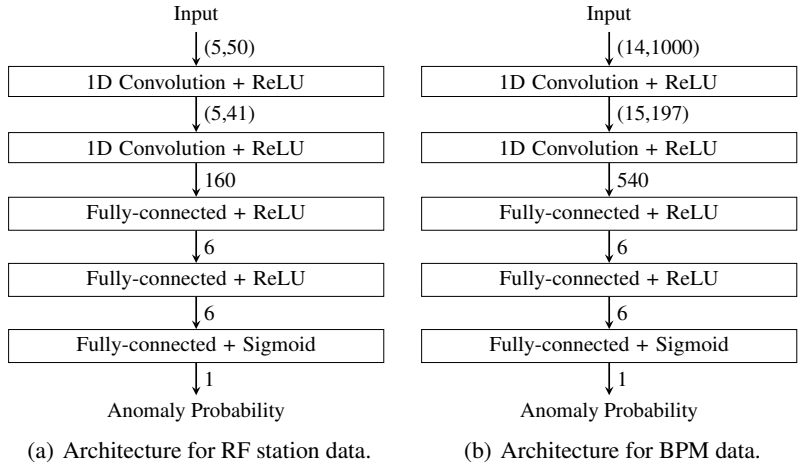
\begin{figure}[htbp]
        \centering
        \subfigure[Architecture for RF station data.]{
            \scalebox{0.8}{
                \begin{tikzpicture}[node distance=1.1cm]
                    \node (data_in) [] {Input};
                    \node (conv1) [nn, below of=data_in] {1D Convolution + ReLU};
                    \node (conv2) [nn, below of=conv1] {1D Convolution + ReLU};
                    \node (fc1) [nn, below of=conv2] {Fully-connected + ReLU};
                    \node (fc2) [nn, below of=fc1] {Fully-connected + ReLU};
                    \node (fc3) [nn, below of=fc2] {Fully-connected + Sigmoid};
                    \node (pred) [below of=fc3] {Anomaly Probability};
            
                    \draw [arrow] (data_in) -- (conv1) node [pos=0.5, right] {(5,50)};
                    \draw [arrow] (conv1) -- (conv2) node [pos=0.5, right] {(5,41)};
                    \draw [arrow] (conv2) -- (fc1) node [pos=0.5, right] {160};
                    \draw [arrow] (fc1) -- (fc2) node [pos=0.5, right] {6};
                    \draw [arrow] (fc2) -- (fc3) node [pos=0.5, right] {6};
                    \draw [arrow] (fc3) -- (pred) node [pos=0.5, right] {1};
                \end{tikzpicture}
            }
            \label{fig:rf_arch_s}
        }
        \subfigure[Architecture for BPM data.]{
            \scalebox{0.8}{
                \begin{tikzpicture}[node distance=1.1cm]
                    \node (data_in) [] {Input};
                    \node (conv1) [nn, below of=data_in] {1D Convolution + ReLU};
                    \node (conv2) [nn, below of=conv1] {1D Convolution + ReLU};
                    \node (fc1) [nn, below of=conv2] {Fully-connected + ReLU};
                    \node (fc2) [nn, below of=fc1] {Fully-connected + ReLU};
                    \node (fc3) [nn, below of=fc2] {Fully-connected + Sigmoid};
                    \node (pred) [below of=fc3] {Anomaly Probability};
    
                    \draw [arrow] (data_in) -- (conv1) node [pos=0.5, right] {(14,1000)};
                    \draw [arrow] (conv1) -- (conv2) node [pos=0.5, right] {(15,197)};
                    \draw [arrow] (conv2) -- (fc1) node [pos=0.5, right] {540};
                    \draw [arrow] (fc1) -- (fc2) node [pos=0.5, right] {6};
                    \draw [arrow] (fc2) -- (fc3) node [pos=0.5, right] {6};
                    \draw [arrow] (fc3) -- (pred) node [pos=0.5, right] {1};
                \end{tikzpicture}
            }
            \label{fig:rf_arch_q}
        }
        \caption{\label{fig:rf_arch} Model architecture for the particle accelerator DNNs, showing the input size to each layer.}
    \end{figure}

    For the comparision to OCSVM~\cite{Scholkopf01}, we use the implementation and default parameters in the PyOD library~\cite{zhao2019pyod}, except for setting the contamination to $17\%$. For the comparison to DGHL~\cite{challu2022deep} and OmniAnomaly~\cite{su2019robust}, we leverage their public repositories: \href{https://github.com/cchallu/dghl}{DGHL} and \href{https://github.com/NetManAIOps/OmniAnomaly}{OmniAnomaly}. For DGHL, we used the default training parameters and architecture, except for setting the feature size ($14$ for BPM data, $1$ for RF data, and $15$ for stacked data), window size for RF data to $16$, and learning rate to $10^{-4}$. For OmniAnomaly, we used the default training parameters and architecture, except for setting the window length to $50$. Due to an issue with the public repository, we ran on a reduce random subset of the training data to fit within GPU memory and avoid OOM errors.

\end{toappendix}

\subsection{Limitations}
The main limitation of this work is the assumption that we have two separate inputs $s$ and $q$, which requires slicing the full set of input features into two groups that are separately indicative of failures. In the course of deriving our method and the theoretical results, we also relied on the assumption that the inputs $s$ and $q$ are independent given the true label. This condition is a sufficient but not necessary condition for our analysis; in practice, including the two real-world data sets shown here, we find our method to work on a broader set of cases than this sufficient condition implies. 




\section{Conclusion}

This paper introduced a new unsupervised approach for anomaly detection that relies on detecting coincident anomalies in two data inputs. By driving two anomaly detection models, one for each set of data, to agree with each other, we eliminate the need for labels. We derive several theoretical properties of our metric $\hat{F}_\beta$, revealing it as a type of clustering algorithm whose behavior is configurable with $\beta$.
Although we only consider a single value of $\beta$ when optimizing $\hat{F}_\beta$ in this work, our approach can be generalized to optimize an entire frontier of choices concurrently, allowing for the model backbones to be reused. Lastly, we show that our method achieves performance levels close to supervised methods on a variety of data sets, including synthetic and real, and time-series and image-based.

\bibliographystyle{ieeetr}
\bibliography{CoincidentAD}

\newpage
\appendix

\end{document}